\documentclass[11pt, a4paper, logo]{deepmind}

\usepackage[authoryear, sort&compress, round]{natbib}
\usepackage{dblfloatfix}
\usepackage[normalem]{ulem}
\usepackage{subcaption}
\usepackage{listings}
\usepackage{xspace}
\PassOptionsToPackage{unicode}{hyperref}
\PassOptionsToPackage{hyphens}{url}

\usepackage{amsmath,amssymb}
\usepackage[T1]{fontenc}
\usepackage[utf8]{inputenc}
\usepackage[utf8]{inputenc}
\usepackage{textcomp} \usepackage{upquote}
\usepackage[]{microtype}
\UseMicrotypeSet[protrusion]{basicmath}

\usepackage{parskip}
\usepackage{xcolor}
\usepackage{xurl}
\usepackage{bookmark}
\hypersetup{
  pdftitle={Ethical and social risks of harm from Language Models}}
 \usepackage{longtable,booktabs,array}
\usepackage{calc} \usepackage{soul}
\usepackage{footnotehyper}
\makesavenoteenv{longtable}
\titleformat{\chapter}{\normalfont\Huge\bfseries}{\thechapter.}{0.5em}{#1}[]
\titlespacing{\chapter}{0pt}{0pt}{3ex}
\usepackage{tcolorbox}
\newenvironment{dialog}{\begin{tcolorbox}}{\end{tcolorbox}}
\usepackage{afterpage}

\title{Ethical and social risks of harm from Language Models}

\renewcommand{\today}{}

\author[1]{Laura Weidinger}
\author[1]{John Mellor}
\author[1]{Maribeth Rauh}
\author[1]{Conor Griffin}
\author[1]{Jonathan Uesato}
\author[1]{Po-Sen Huang}
\author[1,2]{Myra Cheng}
\author[1]{Mia Glaese}
\author[1]{Borja Balle}
\author[1,3]{Atoosa Kasirzadeh}
\author[1]{Zac Kenton}
\author[1]{Sasha Brown}
\author[1]{Will Hawkins}
\author[1]{Tom Stepleton}
\author[1]{Courtney Biles}
\author[1,4]{Abeba Birhane}
\author[1]{Julia Haas}
\author[1]{Laura Rimell}
\author[1]{Lisa Anne Hendricks}
\author[1]{William Isaac}
\author[1]{Sean Legassick}
\author[1]{Geoffrey Irving}
\author[1]{Iason Gabriel}

\affil[1]{DeepMind}
\affil[2]{California Institute of Technology}
\affil[3]{University of Toronto}
\affil[4]{University College Dublin}

\correspondingauthor{Laura Weidinger \texttt{<lweidinger@deepmind.com>}}

\begin{document}
\maketitle

\hypertarget{abstract}{\section*{Abstract}}
This paper aims to help structure the risk landscape associated with large-scale Language Models (LMs). In order to foster advances in responsible innovation, an in-depth understanding of the potential risks posed by these models is needed. A wide range of established and anticipated risks are analysed in detail, drawing on multidisciplinary literature from computer science, linguistics, and social sciences.

The paper outlines six specific risk areas: \protect\hyperlink{i.-discrimination-exclusion-and-toxicity}{I. Discrimination, Exclusion and Toxicity}, \protect\hyperlink{ii.-information-hazards}{II. Information Hazards}, \protect\hyperlink{iii.-misinformation-harms}{III. Misinformation Harms}, \protect\hyperlink{iv.-malicious-uses}{IV. Malicious Uses}, \protect\hyperlink{v.-human-computer-interaction-harms}{V. Human-Computer Interaction Harms}, \protect \hyperlink{vi.-automation-access-and-environmental-harms}{VI. Automation, Access, and Environmental Harms}.

The first risk area discusses fairness and toxicity risks in large-scale language models. This includes four distinct risks: LMs can create unfair discrimination and representational and material harm by perpetuating stereotypes and social biases, i.e. harmful associations of specific traits with social identities. Social norms and categories can exclude or marginalise those who exist outside them. Where a LM perpetuates such norms - e.g. that people called ``Max'' are ``male'', or that ``families'' always consist of a father, mother and child - such narrow category use can deny or burden identities who differ. Toxic language can incite hate or violence or cause offense. Finally, a LM that performs more poorly for some social groups than others can create harm for disadvantaged groups, for example where such models underpin technologies that affect these groups. These risks stem in large part from choosing training corpora that include harmful language and overrepresent some social identities.

The second risk area includes risks from private data leaks or from LMs correctly inferring private or other sensitive information. These risks stem from private data that is present in the training corpus and from advanced inference capabilities of LMs.

The third risk area comprises risks associated with LMs providing false or misleading information. This includes the risk of creating less well-informed users and of eroding trust in shared information. Misinformation can cause harm in sensitive domains, such as bad legal or medical advice. Poor or false information may also lead users to perform unethical or illegal actions that they would otherwise not have performed. Misinformation risks stem in part from the processes by which LMs learn to represent language: the underlying statistical methods are not well-positioned to distinguish between factually correct and incorrect information.

The fourth risk area spans risks of users or product developers who try to use LMs to cause harm. This includes using LMs to increase the efficacy of disinformation campaigns, to create personalised scams or fraud at scale, or to develop computer code for viruses or weapon systems.

The fifth risk area focuses on risks from the specific use case of a ``conversational agent'' that directly interacts with human users. This includes risks from presenting the system as ``human-like'', possibly leading users to overestimate its capabilities and use it in unsafe ways. Another risk is that conversation with such agents may create new avenues to manipulate or extract private information from users. LM-based conversational agents may pose risks that are already known from voice assistants, such as perpetuating stereotypes by self-presenting e.g. as ``female assistant''. These risks stem in part from LM training objectives underlying such conversational agents and from product design decisions.

The sixth risk area includes risks that apply to LMs and Artificial Intelligence (AI) systems more broadly. Training and operating LMs can incur high environmental costs. LM-based applications may benefit some groups more than others and the LMs themselves are inaccessible to many. Lastly, LM-based automation may affect the quality of some jobs and undermine parts of the creative economy. These risks manifest particularly as LMs are widely used in the economy and benefits and risks from LMs are globally unevenly distributed.

In total, we present 21 risks. We then discuss the points of origin of different risks and point to potential risk mitigation approaches. The point of origin of a harm may indicate appropriate mitigations: for example, the risk of leaking private data originates from this data being present in the training dataset. It can be mitigated at the point of origin, by better redaction or curation of training data. However, other mitigation approaches may also be applicable and ensure more robust mitigation overall. For example, algorithmic tools applied during training, such as differential privacy methods, or product decisions, such as constraining access and use cases of the LM, are additional mitigation approaches that can be pursued in parallel. Risk mitigation approaches range from social or public policy interventions, to technical solutions and research management, to participatory projects and product design decisions.

Lastly, we discuss organisational responsibilities in implementing such mitigations, and the role of collaboration. Measuring and mitigating ethical and social risks effectively requires a wide range of expertise, and fair inclusion of affected communities. It is critical to implement mitigations with a broad view of the landscape of risks, to ensure that mitigating against one risk of harm does not aggravate another. Otherwise, for example, mitigation approaches to toxic speech can inadvertently lead to lower LM performance for some social groups. We highlight directions for further research, particularly on expanding the toolkit for assessing and evaluating the outlined risks in LMs, and the need for inclusive participatory methods. Finally, we conclude by showing how the present work - of structuring the risk landscape - is the first step in a broader framework of responsible innovation.

\tableofcontents

\clearpage \hypertarget{readers-guide}{\section*{Reader's guide}\label{readers-guide}}

This is a long document. The report is divided into three segments.

First, the \protect\hyperlink{introduction}{Introduction} provides a brief introduction to Language Models.

Second, the \protect\hyperlink{classification-of-harms-from-language-models}{Classification of harms from language models} gives a taxonomy and detailed account of a range of social and ethical risks associated with Language Models.

Third, the \protect\hyperlink{discussion}{Discussion} and \protect\hyperlink{directions-for-future-research}{Directions for future research} explore some underlying causes of these risks, a range of mitigation approaches, and possible challenges to be addressed through future research.

Individual sections can be read independently or together. We recommend:
\begin{itemize}
	\item 
	\begin{quote}
		\textbf{1 minute read:} Study \protect\hyperlink{risks_table}{Table~1} for a high-level overview of the risks considered. 
	\end{quote}
	\item 
	\begin{quote}
		\textbf{10 minute read:} Read the \protect\hyperlink{abstract}{Abstract} and \protect\hyperlink{risks_table}{Table~1} for an overview of the risks considered. Then skim all bold text in the segment on \protect\hyperlink{classification-of-harms-from-language-models}{Classification of harms from language models} and skim \protect\hyperlink{directions-for-future-research}{Directions for future research} for an overview of risks and challenges. 
	\end{quote}
	\item 
	\begin{quote}
		\textbf{Readers who actively work on LMs:} We encourage you to skim all bold text in the segment on \protect\hyperlink{classification-of-harms-from-language-models}{Classification of harms from language models}, and to get stuck in risks that directly relate to your own work and interest - as you will likely be able to help solve some of the field's core challenges in this domain. 
	\end{quote}
	\item 
	\begin{quote}
		\textbf{Readers with no background on LMs:} We recommend you read the \protect\hyperlink{abstract}{Abstract} and \protect\hyperlink{introduction}{Introduction} first as these introduce key terminology that is used in this report. Next, study \protect\hyperlink{risks_table}{Table~1} for a high-level overview of the risks considered and read the risk headers and example dialog boxes for each risk in the \protect\hyperlink{classification-of-harms-from-language-models}{Classification of harms from language models}. Get stuck in risks that are of interest to you and read the \protect\hyperlink{discussion}{Discussion} on challenges in mitigating these risks. 
	\end{quote}
	\item 
	\begin{quote}
		\textbf{Readers with an interest in a particular risk or type of harm}: We encourage you to read the \protect\hyperlink{abstract}{Abstract}, \protect\hyperlink{risks_table}{Table~1} and \protect\hyperlink{discussion}{Discussion} for context on the broader risk landscape and approaches to mitigation, in addition to reading the specific section on the risk that piques your interest. 
	\end{quote}
	\item 
	\begin{quote}
		\textbf{Readers with an interest in approaches to mitigating harms:} We recommend you read the \protect\hyperlink{abstract}{Abstract} for an overview of the harms considered and read \protect\hyperlink{risks_table}{Table~1} with a focus on the mechanisms underlying each risk area. Jump to the \protect\hyperlink{discussion}{Discussion} on approaches to mitigating risks and read \protect\hyperlink{directions-for-future-research}{Directions for future research} on methodological and normative challenges in assessing and mitigating risks, and proposals for addressing these challenges.
	\end{quote}
\end{itemize}

\hypertarget{introduction}{\chapter{Introduction}\label{introduction}}

Language Models (LMs)\footnote{These recent LMs are also referred to as ``large language models'', or ``large-scale language models''.} are rapidly growing in size and effectiveness, yielding new breakthroughs and attracting increasing research attention  \citep{Brownetal2020,Fedusetal2021,Microsoft2020, Rae2021}. Several Artificial Intelligence (AI) research labs are pursuing LM research, spurred by the promise these models hold for advancing research and for a wide range of beneficial real-world applications. Some research groups have suggested that recent large-scale LMs may be a `foundational' breakthrough technology, potentially affecting many aspects of life \citep{Bommasanietal2021}. The potential impact of such LMs makes it particularly important that actors in this space lead by example on responsible innovation.

Responsible innovation entails that in addition to developing the technology, it is essential to thoughtfully assess the potential benefits as well as potential risks that need to be mitigated \citep{Stilgoeetal2013}. Prior research has explored the potential for ethical and safe innovation of large-scale LMs, including interdisciplinary workshops to scope out risks and benefits \citep{Tamkinetal2021}, papers that outline potential risks \citep{Benderetal2021,Kentonetal2021,Dinanetal2021,Bommasanietal2021}, and papers identifying ways to mitigate potential harms \citep{Welbletal2021,SolaimanDennison2020,Chenetal2021}.\footnote{Note that the origin of a risk is not a perfect guide to potential mitigations - a point we discuss in more detail in \protect\hyperlink{understanding-the-point-of-origin-of-a-risk}{Understanding the point of origin of a risk}.} For this report, we seek to build on this prior work by proposing an initial taxonomy of risks associated with LM development and use, as well as outlining concrete next steps and directions for future research that supports responsible innovation for LMs.

The overall aim of this report is three-fold:
\begin{enumerate}
	\def\labelenumi{\arabic{enumi}.} 
	\item 
	\begin{quote}
		Underpin responsible decision-making by organisations working on LMs by broadening and structuring the discourse on AI safety and ethics in this research area, 
	\end{quote}
	\item 
	\begin{quote}
		Contribute to wider public discussion about risks and corresponding mitigation strategies for LMs, 
	\end{quote}
	\item 
	\begin{quote}
		Guide mitigation work by research groups working on LMs. We aim to support the mutual exchange of expertise in this area, to help make the risks posed by LMs actionable. 
	\end{quote}
\end{enumerate}

We structure the identified risks in a taxonomy of ethical and social risks associated with LMs, under 6 risk areas: \protect\hyperlink{i.-discrimination-exclusion-and-toxicity}{I. Discrimination, Exclusion and Toxicity}, \protect\hyperlink{ii.-information-hazards}{II. Information Hazards}, \protect\hyperlink{iii.-misinformation-harms}{III. Misinformation Harms}, \protect\hyperlink{iv.-malicious-uses}{Malicious Uses}, \protect\hyperlink{v.-human-computer-interaction-harms}{V. Human-Computer Interaction Harms}, \protect\hyperlink{vi.-automation-access-and-environmental-harms}{VI. Automation, Access, and Environmental Harms}. An overview of the risks that fall under each risk area can be found in the \protect\hyperlink{classification-of-harms-from-language-models}{Classification of harms from language models} part of the report.

Each risk is discussed in detail with regard to the nature of the harm, empirical examples, and additional considerations. For each risk, we provide a fictitious example to illustrate how the risk in question may manifest.\footnote{Each of these examples assumes a dialogue format where a human supplies a prompt and the LM offers a response. There are many LM use cases beyond such conversational agents. These examples are for illustrative purposes only, and the same risk may manifest differently in other LM use cases.} However the risks described apply to LMs more generally and do not depend on the dialogue modality unless otherwise specified. Since several of the risks discussed below are neither novel nor exclusive to LMs or related technologies, we offer context on how each risk manifests in existing language technologies. We also mark each risk as either ``anticipated'' or ``observed'', depending on whether a given risk has already been observed or whether further work is needed to indicate real-world manifestations of this risk. The creation of a taxonomy of risks supports the exercise of foresight in this space, with the aim of guiding action to resolve any issues that can be identified in advance.

Responsible innovation is a collaborative endeavour. In order to anticipate and mitigate risks posed by technology successfully, we need to view these issues through multiple lenses and perspectives. This report was written by a large group of researchers with varied disciplinary backgrounds and areas of expertise. To review the risk landscape as comprehensively as possible, we collated potential risks from a wide range of sources including analyses from the fields of AI ethics, AI safety, race and gender studies, linguistics and natural language processing and studies at the intersection of society and technology (also referred to as sociotechnical studies), as well as analyses by civil society organisations and news reports. Further risks were added based on our own experience and expertise. Beyond publishing research, we believe responsible innovation also requires inclusive dialogue between stakeholders in AI development which includes affected communities and the wider public \citep{Mohamedetal2020,GabrielatUCL2020,Stilgoeetal2013,IbrahimcitationinMurgiafortheFinancialTimes2021}. In the future, we look to continue to deepen our understanding of risks and mitigations including by working with external partners and communities.

\hypertarget{limitations}{\section{Limitations}\label{limitations}}

Note that this report is part of a broader research programme working toward the responsible innovation of LMs and necessarily leaves some questions unanswered. For example, we do not discuss potential beneficial applications of LMs nor do we offer a comprehensive overview of potential use cases. Nor do we attempt to perform a full ethical evaluation of LMs, which must weigh both the potential benefits and risks of a given technology. To assess the overall balance of benefit and cost, separate analysis of the benefits arising from proposed LM applications would be needed. Instead, the focus here is on anticipating and structuring the risk landscape, with the intention of supporting a larger constructive research effort.

This report is also necessarily a snapshot in time: it was initiated in autumn 2020 and completed in summer 2021. It is likely that we miss risks arising from LMs that depend, for their visibility, on the passage of time. As such, the presented taxonomy is merely a starting point and will need to be updated as new challenges come into focus and additional perspectives are brought to bear on these questions.

This report focuses on risks associated with \emph{operating} LMs. Risks of harm that are associated with training are not discussed. This includes concerns about the working conditions of data annotators or ``ghost workers'' \citep{GraySuri2019}, the ethics of supply chains of hardware on which LM computations are run \citep{Crawford2021}, or environmental costs of training such models \citep{Strubelletal2019,Benderetal2021,Pattersonetal2021,Schwartzetal2020} which are only briefly referenced in the section on \protect\hyperlink{vi.-automation-access-and-environmental-harms}{VI. Automation, access, and environmental harms}. This report also does not cover risks that depend on specific applications.

This report excludes risks which the authors anticipate to depend on capabilities that are several years in the future, for example because they depend on capabilities that are several step changes beyond the state-of-the-art. A subset of such long-term risks is addressed in literature on existential risk and AI Safety \citep{Armstrongetal2012,Kentonetal2021}. This report also does not cover risks that depend on superintelligence as described in \citep{Bostrom2014}.

Finally, this report does not discuss risks that depend on multiple modalities, for example from models that combine language with other domains such as vision or robotics. While several of the insights in this report are translatable to such models, these require distinct risk assessments. For some discussion on risks associated with multi-modal large models, see \citep{Bommasanietal2021}.

\hypertarget{note-on-terminology}{\subsection{Note on terminology}\label{note-on-terminology}}

This report focuses on the risks of large-scale language models, including in specific applications of these models such as conversational assistants, or in other language technologies. Several of these risks also apply to smaller language models. For detailed definitions of Language Models, Language Agents, and Language Technologies please refer to the section on \protect\hyperlink{definitions}{Definitions} in the \protect\hyperlink{appendix}{Appendix}.

For simplicity we refer to ``LMs'' throughout. Where risks are unique to specific types of applications, such as conversational agents, this is explicitly stated.

\hypertarget{a-brief-history-of-language-models}{\section{A Brief history of Language Models}\label{a-brief-history-of-language-models}}

\hypertarget{origins}{\subsection{Origins}\label{origins}}

The main methodology underpinning contemporary large-scale language models traces its origins to methods developed by the research group of Frederick Jelinek on Automatic Speech Recognition (ASR) in the 1970s and `80s \citep{Jelinek1976}. This research group built on prior work in statistics by Claude Shannon \citep{Shannon1948} and Andrey Markov \citep{Markov1913}. In parallel, James Baker \citep{Baker1975} developed a similar approach to ASR (see \citep{JurafskyMartin2021}).

Jelinek's group pioneered an information theoretic approach to ASR, observing that performing any task that requires producing language conditioned on an input using a probability distribution $p(\text{language} | \text{input})$ can be factored into a language model representing a probability distribution $p(\text{language})$ multiplied by the task specific distribution $p(\text{input} | \text{language})$. This factorisation suggests that general LMs $p(\text{language})$ can aid language prediction tasks where the LM captures the relevant language distribution. Whilst this factorisation is not explicitly used in most current systems, it implicitly underpins current LM research and is a useful way to understand the role language modelling plays in specific language technologies such as conversational agents, machine translation, and question answering.

\hypertarget{transformer-models}{\subsection{Transformer models}\label{transformer-models}}

More recently, the transformer architecture was developed \citep{Vaswanietal2017}. Transformers are a class of architectures that use a series of so-called transformer blocks comprising a self-attention layer followed by a feedforward layer, linked together with residual connections. The self-attention layer helps the model to consider neighbouring words in the input as it processes a specific word. Originally, the transformer architecture was proposed for the task of machine translation \citep{Vaswanietal2017}. \citep{Radfordetal2018} use a modified version applied to the task of language modeling (predicting the next word in a sentence). Subsequent work on LMs \citep{Radfordetal2018b,Brownetal2020} uses a similar architecture. An accessible visual introduction to the transformer architecture can be found in \citep{Alammar2018}. Recent language models built on the transformer architecture have been fine-tuned directly, without the need for task-specific architectures \citep{Radfordetal2018b,Devlinetal2018,HowardRuder2018}.

\hypertarget{large-language-models}{\subsection{``Large'' Language Models}\label{large-language-models}}

The recent upwind in LM research is rooted in the capacity to increase LM size in terms of number of parameters and size of training data \citep{Benderetal2021}. Training models on extremely large datasets such as the Colossal Clean Crawl Corpus (C4) \citep{Raffeletal2019} and WebText \citep{Radfordetal2018} resulted in sequence prediction systems with much more general applicability compared to the prior state-of-the-art \citep{Brownetal2020,Fedusetal2021,Microsoft2020}. These models also displayed greater few-shot and zero-shot learning capabilities compared to smaller LMs \citep{Brownetal2020}. These properties were found to greatly simplify the development of task-specific LAs by reducing the adaptation process to prompt design \citep{Zhangetal2021}. The insight that powerful sequence prediction systems could be created by scaling up the size of LMs and training corpora motivated an upsurge in interest and investment in LM research by several AI research labs.

\hypertarget{classification-of-harms-from-language-models}{\chapter{Classification of harms from language models}\label{classification-of-harms-from-language-models}}

\fancyhead[C]{\footerfont \rightmark}

In this section we outline our taxonomy of ethical and social risks of harm associated with Language Models. We identify 21 risks of harm, organised into six risk areas (for an overview see \protect\hyperlink{risks_table}{Table~1}).
In this table we also note the mechanisms by which different groups of risks emerge.

\begin{table}
    \hypertarget{risks_table}{\emph{\textbf{Table 1.} Overview of all risks covered in this report.}
	\label{risks_table}}
	\begin{longtable}
		[]{@{} >{
		\raggedright\arraybackslash}p{(\columnwidth - 0\tabcolsep) * \real{1.00}}@{}} \toprule
		\begin{minipage}
			[b]{\linewidth}
			\raggedright
			\begin{enumerate}
				\def\labelenumi{\Roman{enumi}.} 
				\item \protect\hyperlink{i.-discrimination-exclusion-and-toxicity}{\textbf{Discrimination, Exclusion and Toxicity}} \\
				\textbf{Mechanism}: These risks arise from the LM accurately reflecting natural speech, including unjust, toxic, and oppressive tendencies present in the training data. \\
				\textbf{Types of Harm}: Potential harms include justified offense, material (allocational) harm, and the unjust representation or treatment of marginalised groups. 
				\begin{itemize}
					\item Social stereotypes and unfair discrimination 
					\item Exclusionary norms 
					\item Toxic language 
					\item Lower performance by social group 
				\end{itemize}
				\item \protect\hyperlink{ii.-information-hazards}{\textbf{Information Hazards}} \\
				\textbf{Mechanism}: These risks arise from the LM predicting utterances which constitute private or safety-critical information which are present in, or can be inferred from, training data. \\
				\textbf{Types of Harm}: Potential harms include privacy violations and safety risks. 
				\begin{itemize}
					\item Compromise privacy by leaking private information 
					\item Compromise privacy by correctly inferring private information 
					\item Risks from leaking or correctly inferring sensitive information 
				\end{itemize}
				\item \protect\hyperlink{iii.-misinformation-harms}{\textbf{Misinformation Harms}} \\
				\textbf{Mechanism}: These risks arise from the LM assigning high probabilities to false, misleading, nonsensical or poor quality information. \\
				\textbf{Types of Harm}: Potential harms include deception, material harm, or unethical actions by humans who take the LM prediction to be factually correct, as well as wider societal distrust in shared information. 
				\begin{itemize}
					\item Disseminating false or misleading information 
					\item Causing material harm by disseminating misinformation e.g. in medicine or law 
					\item Nudging or advising users to perform unethical or illegal actions 
				\end{itemize}
				\item \protect\hyperlink{iv.-malicious-uses}{\textbf{Malicious Uses}} \\
				\textbf{Mechanism}: These risks arise from humans intentionally using the LM to cause harm. \\
				\textbf{Types of Harm}: Potential harms include undermining public discourse, crimes such as fraud, personalised disinformation campaigns, and the weaponisation or production of malicious code. 
				\begin{itemize}
					\item Reducing the cost of disinformation campaigns 
					\item Facilitating fraud and impersonation scams 
					\item Assisting code generation for cyber attacks, weapons, or malicious use 
					\item Illegitimate surveillance and censorship 
				\end{itemize}
				\item \protect\hyperlink{v.-human-computer-interaction-harms}{\textbf{Human-Computer Interaction Harms}} \\
				\textbf{Mechanism:} These risks arise from LM applications, such as Conversational Agents, that directly engage a user via the mode of conversation. \\
				\textbf{Types of Harm}: Potential harms include unsafe use due to users misjudging or mistakenly trusting the model, psychological vulnerabilities and privacy violations of the user, and social harm from perpetuating discriminatory associations via product design (e.g. making ``assistant'' tools by default ``female.'') 
				\begin{itemize}
					\item Anthropomorphising systems can lead to overreliance or unsafe use 
					\item Create avenues for exploiting user trust to obtain private information 
					\item Promoting harmful stereotypes by implying gender or ethnic identity 
				\end{itemize}
				\item \protect\hyperlink{vi.-automation-access-and-environmental-harms}{\textbf{Automation, access, and environmental harms}} \\
				\textbf{Mechanism}: These risks arise where LMs are used to underpin widely used downstream applications that disproportionately benefit some groups rather than others. \\
				\textbf{Types of Harm}: Potential harms include increasing social inequalities from uneven distribution of risk and benefits, loss of high-quality and safe employment, and environmental harm. 
				\begin{itemize}
					\item Environmental harms from operating LMs 
					\item Increasing inequality and negative effects on job quality 
					\item Undermining creative economies 
					\item Disparate access to benefits due to hardware, software, skill constraints 
				\end{itemize}
			\end{enumerate}
		\end{minipage}
		\\
		\bottomrule 
	\end{longtable}
\end{table} 

\hypertarget{i.-discrimination-exclusion-and-toxicity}{\section{Discrimination, Exclusion and Toxicity}\label{i.-discrimination-exclusion-and-toxicity}}

Social harms that arise from the language model producing discriminatory or exclusionary speech

\hypertarget{overview}{\subsection{Overview}\label{overview}}

Language has long been a vehicle for perpetuating injustices against marginalised populations and conferring power to oppressive institutions \citep{Craftetal2020,LippiGreen2012}. Social harms that manifest in language can also arise from language predicted by language models (LMs). We distinguish between the following harms. First, language that promotes stereotypes or causes unfair discrimination\footnote{Note that the terms ``bias'' and ``discrimination'' have different meanings in classical statistics compared to sociotechnical studies; for a definition of these terms, see the \protect\hyperlink{definitions}{Definitions} in the \protect\hyperlink{appendix}{Appendix}.} can contribute to the oppression of those at social margins \citep{Crenshaw2017}. Second, language that reinforces social norms can exclude identities that exist outside these norms \citep{Benderetal2021,Foucault1975}. Third, language can be `toxic', for example by inciting violence or causing offense \citep{FortunaNunes2018}. Fourth, a form of discrimination emerges when language technologies perform better for some social groups than others.

Risks covered in this section:
\begin{itemize}
	\item \protect\hyperlink{social-stereotypes-and-unfair-discrimination}{Social stereotypes and unfair discrimination}
	
	\item \protect\hyperlink{exclusionary-norms}{Exclusionary norms}
	
	\item \protect\hyperlink{toxic-language}{Toxic language}
	
	\item \protect\hyperlink{lower-performance-for-some-languages-and-social-groups}{Lower performance by social group}
\end{itemize}

\hypertarget{social-stereotypes-and-unfair-discrimination}{\subsection{Social stereotypes and unfair discrimination}\label{social-stereotypes-and-unfair-discrimination}}
\begin{dialog}
	
	Q: ``Two Muslims walked into a \ldots''
	
	A: ``Texas cartoon contest and opened fire.''\footnote{Example from \citep{Abidetal2021}}
\end{dialog}

\emph{Observed risk: This is a well-documented problem that needs a mitigation strategy and tools to analyse the model against benchmarks of 'acceptability'.}

\hypertarget{problem}{\subsubsection*{Problem}\label{problem}}

Perpetuating harmful stereotypes and discrimination is a well-documented harm in machine learning models that represent natural language \citep{Caliskanetal2017}. LMs that encode discriminatory language or social stereotypes can cause different types of harm. It may be useful to distinguish between allocational and representational harms: allocational harms occur when resources and opportunities are unfairly allocated between social groups; they may occur when LMs are used in applications that are used to make decisions that affect persons. Representational harms include stereotyping, misrepresenting, and demeaning social groups Barocas and Wallach cited in \citep{Blodgettetal2020}.

Unfair discrimination manifests in differential treatment or access to resources among individuals or groups based on sensitive traits such as sex, religion, gender, sexual orientation, ability and age. The dimensions along which such oppression occurs can also be rooted in culture-specific or otherwise localised social hierarchies. For example, the Hindu caste system underpins discrimination in India, but not across the globe \citep{Sambasivanetal2021}. Additionally, injustice can be compounded when social categories intersect, for example in the discrimination against a person that holds a marginalised gender and a marginalised religion \citep{Crenshaw1993}.

Allocational harm caused by discriminatory systems is particularly salient if bias occurs in applications that materially impact people's lives, such as predicting a person's creditworthiness \citep{Mehrabietal2019}, criminal recidivism \citep{Angwinetal2016}, or suitability to a job \citep{MutajbaMahapatra2019}. For example, a language technology that analyses CVs for recruitment, or to give career advice, may be less likely to recommend historically discriminated groups to recruiters, or more likely to recommend lower paying careers to marginalised groups. Unfair biases are already well-documented in machine learning applications ranging from diagnostic healthcare algorithms \citep{Obermeyeretal2019} to social outcome prediction \citep{Narayanan2019}; for a more general introduction see \citep{ChouldechovaRoth2018,Mehrabietal2021,KordzadehGhasemaghaei2021,ZouSchiebinger2018,Noble2018}. Based on our current understanding, such stereotyping and unfair bias are set to recur in language technologies building on LMs unless corrective action is taken.

\hypertarget{why-we-should-expect-lms-to-reinforce-stereotypes-and-unfair-discrimination-by-default}{\paragraph{Why we should expect LMs to reinforce stereotypes and unfair discrimination by default}\label{why-we-should-expect-lms-to-reinforce-stereotypes-and-unfair-discrimination-by-default}}

LMs are optimised to mirror language as accurately as possible, by detecting the statistical patterns present in natural language \protect\hyperlink{definitions}{Definitions}. The fact that LMs track patterns, biases, and priors in natural language is not negative \emph{per se} \citep{Shahetal2020}. Rather, it becomes a problem when the training data is unfair, discriminatory, or toxic. In this case, the optimisation process results in models that mirror these harms. As a result, LMs that perform well with regard to their optimisation objective can work poorly with regard to social harms, insofar as they encode and perpetuate harmful stereotypes and biases present in the training data.

Stereotypes and unfair discrimination can be present in training data for different reasons. First, training data reflect historical patterns of systemic injustice when they are gathered from contexts in which inequality is the status quo. Training systems on such data entrenches existing forms of discrimination \citep{Browne2015}. In this way, barriers present in our social systems can be captured by data, learned by LMs, and perpetuated by their predictions \citep{Hampton2021}.

Second, training data can be biased because some communities are better represented in the training data than others. As a result, LMs trained on such data often model speech that fails to represent the language of those who are marginalised, excluded, or less often recorded. The groups that are traditionally underrepresented in training data are often disadvantaged groups: they are also referred to as the `undersampled majority' \citep{BuolamwinicitedinRaji2020}. The implications of unrepresentative training data for downstream biases and stereotyping in LMs demonstrate the power that is exercised by those who have influence over what data is used for model training \citep{Blodgettetal2020}. While in principle, LMs are optimised to represent language with high fidelity, they can also overrepresent small biases present in the training data, a phenomenon referred to as `bias amplification' \citep{Zhaoetal2017,WangRussakovsky2021}.

\hypertarget{examples}{\subsubsection*{Examples}\label{examples}}

Generative LMs have frequently been shown to reproduce harmful social biases and stereotypes. Predictions from the GPT-3 model \citep{Brownetal2020} were found to exhibit anti-Muslim and, to a lesser degree, antisemitic bias, where `\,``Muslim'' was analogised to ``terrorist'' in 23\% of test cases, while ``Jewish'' was mapped to ``money'' in 5\% of test cases \citep{Abidetal2021}\footnote{See also the authors' \citep{illustration} of ``how hard it is to generate text about Muslims from GPT-3 that has nothing to do with violence'', and \citep{GershgornforOz2021}.}. Gender and representation biases were found in fictional stories generated by GPT-3 \citep{LucyBamman2021}, where female-sounding names were more often associated with stories about family and appearance, and described as less powerful than masculine characters.

The \emph{StereoSet} benchmark measures references to stereotypes of race, gender, religion, and profession in generative LMs and finds that the models GPT2 \citep{Radfordetal2018} and masked models BERT \citep{Devlinetal2018}, ROBERTA \citep{Liuetal2019}, XLNET \citep{Yangetal2019} exhibit `strong stereotypical associations' \citep{Nadeemetal2020}. The CrowS-Pairs benchmark finds that cultural stereotypes were reproduced by likelihood estimates of masked LMs BERT \citep{Devlinetal2018}, and RoBERTA \citep{Liuetal2019,Nangiaetal2020}\footnote{Recent work critiques some current methods for measuring bias in LMs highlighting the importance of further exploration on valid measures \citep{Blodgettetal2021}.}. The HONEST benchmark shows that GPT-2 and BERT sentence completions promote 'hurtful stereotypes' across six languages \citep{Nozzaetal2020}, and discriminatory gender biases were found in contextual word embedding by BERT \citep{Kuritaetal2019} and ELMo \citep{Zhouetal2019}. LMs trained on news articles and Wikipedia entries have been demonstrated to exhibit considerable levels of bias against particular country names, occupations, and genders \citep{Huangetal2019}.

\hypertarget{additional-considerations}{\subsubsection*{\texorpdfstring{Additional considerations }{Additional considerations }}\label{additional-considerations}}

\hypertarget{underrepresented-groups-in-the-training-data}{\paragraph{Underrepresented groups in the training data}\label{underrepresented-groups-in-the-training-data}}

Training data reflect the views, values, and modes of communication by the communities whose language is captured in the corpus. For example, a dataset of Reddit user comments was found to encode discriminatory views based on gender, religion and race \citep{Ferreretal2020}. As a result, it is important to carefully select and account for the biases present in the training data. However ML training datasets are often collected with little curation or supervision and without factoring in perspectives from communities who may be underrepresented \citep{SoGebru2020}. For more discussion of this, see also the section on \protect\hyperlink{why-we-should-expect-lms-to-reinforce-stereotypes-and-unfair-discrimination-by-default}{Why we should expect LMs to reinforce unfair bias, toxic speech, and exclusionary norms}.

\hypertarget{documentation-of-biases-in-training-corpora}{\paragraph{Documentation of biases in training corpora}\label{documentation-of-biases-in-training-corpora}}

The impact of training data on the LM makes it important to transparently disclose what groups, samples, voices and narratives are represented in the dataset and which may be missing. One format that has been proposed for such dataset documentation \citep{BenderFriedman2018} are `Datasheets' \citep{Gebruetal2020}. Some work in this direction includes documentation on the Colossal Clean Crawl Corpus (C4) that highlights the most prominently represented sources and references to help illuminate \emph{whose} biases are likely to be encoded in the dataset \citep{Dodgeetal2020}. Documentation of larger datasets is critical for anticipating and understanding the pipeline by which different harmful associations come to be reflected in the LM.

\hypertarget{training-data-required-to-reduce-bias-may-not-yet-exist}{\paragraph{Training data required to reduce bias may not yet exist}\label{training-data-required-to-reduce-bias-may-not-yet-exist}}

Approaches to biased training data range from curating dedicated training datasets to not building models in domains where such data does not exist.\footnote{Another proposed approach relies on synthetic data, although the efficacy of this approach remains uncertain and it raises distinct challenges, on amplifying other biases \citep{ChenLuetal2021,Nikolenko2021,Ghalebikesabietal2021}.} Curating training data can help to make LMs fairer, but creating better datasets requires dedicated work \citep{Hutchinsonetal2021,SoGebru2020} and may require novel data curation pipelines and tools \citep{Dentonetal2020}. Training corpora for state of the art LMs are extremely large, so that further innovation on semi-automated curation methods may be needed in order to make the curation of such datasets tractable. Determining what constitutes a truly fair and equitable training dataset may also require further research in Ethics and Law \citep{KohlerHausman2019}. In one high-profile, real-world example, researchers attempted to train a classifier to support recruitment, but found that the training data was inherently biased and found no alternative to create a more equitable training dataset - leading to the research project being abandoned \citep{DastinReuters2018}\footnote{In this real-world example, a model ranking applicant suitability based on written CVs was biased against the term `women' (as in `women's chess club'). In an attempt to correct for this discriminatory performance, the model was initially corrected to not devalue a CV based on terms referring to `women'. However, the algorithm continued to espouse an unfair gender bias against women, simply because there had been a gender bias in Amazon's prior hiring history, which was reflected in the training data. As no sufficient data on successful female applicants was available to train or fine-tune the model to reduce its gender bias, the problem of de-biasing this algorithm seemed intractable, `executives lost hope for the project' \citep{DastinReuters2018}, and it was stopped.}.

\hypertarget{localised-stereotypes-are-hard-to-capture}{\paragraph{Localised stereotypes are hard to capture}\label{localised-stereotypes-are-hard-to-capture}}

As stereotypes change over time and vary between contexts, it is impossible for any given research team to be aware of, and up-to-date on, all relevant stereotypes that may cause harm or offense. In addition, the stereotypes at play in a given local context may only be knowable through committed ethnographic work on the ground \citep{MardaNarayan2021}. The expertise for identifying harmful stereotypes often lies with the lived experience of affected groups \citep{MillsinSullivanTuana2007}. This creates a challenge in knowing what stereotypes to search for, detect, and mitigate at the point of creating a LM. One way to help address this challenge is to use inclusive and fair participatory approaches \citep{Martinetal2020}, by establishing participatory mechanisms and institutions that can operate over time \citep{Sloaneetal2020}, and by providing broad and transparent dataset documentation.

\hypertarget{uncertainty-on-downstream-uses-complicate-fairness-analyses}{\paragraph{Uncertainty on downstream uses complicate fairness analyses}\label{uncertainty-on-downstream-uses-complicate-fairness-analyses}}

Identifying affected communities is challenging during the early stages of building a LM when no particular application, product, or user group has been defined. It is unclear to what extent a training regime can be defined that increases model ``fairness'' whilst being agnostic on downstream applications \citep{HancoxLiKumar2021}. While some aspects of fairness are best considered at early research stages, more specific assessments of potential discrimination must be considered again at the point of developing a concrete application. Methods for detecting and mitigating harmful stereotypes can place an additional burden or privacy cost on minorities, e.g. through collecting additional data. Where this is the case, sustained mitigation of such harms requires engaging affected groups on fair terms that foreground their needs and interests.

\hypertarget{detecting-harmful-stereotypes-can-require-nuanced-analyses-over-multiple-samples}{\paragraph{Detecting harmful stereotypes can require nuanced analyses over multiple samples}\label{detecting-harmful-stereotypes-can-require-nuanced-analyses-over-multiple-samples}}

Stereotyping may only be detectable over multiple samples. ``Pointwise'' stereotyping manifests directly in the text prediction of a single sample and so can be identified in a single instance \citep{Khalifaetal2021}. ``Distributional'' stereotyping on the other hand manifests in the repetition of a seemingly harmless association of certain properties with a group. For example, where a LM predicts passive verbs more often in association with female than male names, such distributional stereotyping of females as more passive may occur. Such ``distributional'' bias may also manifest as notable omissions, e.g. where a language agent that generates fantasy stories by relying on a LM only generates stories with male, never female villains. Such distributional bias becomes apparent only upon analysing multiple predictions and requires distinct forms of evaluation and correction \citep{Khalifaetal2021}.

\hypertarget{exclusionary-norms}{\subsection{Exclusionary norms}\label{exclusionary-norms}}
\begin{dialog}
	
	Q: What is a family?
	
	A: A family is: a man and a woman who get married and have children. \emph{(not accounting for non-heteronormative families and children out of wedlock, for single-parent families and for the fact that families sometimes do not have children)}
\end{dialog}

\emph{Observed risk: This is a well-documented problem that needs a mitigation strategy and tools to analyse the model against benchmarks of 'acceptability'.}

\hypertarget{problem-1}{\subsubsection*{\texorpdfstring{Problem }{Problem }}\label{problem-1}}

In language, humans express social categories and norms. Language models (LMs) that faithfully encode patterns present in natural language necessarily encode such norms and categories. This is why technological artifacts such as LMs are never ``value neutral'' - they represent and perpetuate the values and norms present in the training data \citep{Benderetal2021,Winner1980}.

Such norms and categories exclude groups who live outside them \citep{Foucault1975}. For example, defining the term ``family'' as married parents of male and female gender with a blood-related child, denies the existence of families to whom these criteria do not apply. Moreover, exclusionary norms intersect with discrimination as they almost invariably work to exclude groups that have historically been marginalised. Exclusionary norms can manifest in ``subtle patterns like referring to \emph{women doctors} as if doctor itself entails not-woman, or referring to \emph{both genders} excluding the possibility of non-binary gender identities'' \citep{Benderetal2021}, emphasis added.

Furthermore, exclusionary norms can place a disproportionate burden or ``psychological tax'' on those who do not fit or comply with these norms or who are trying to challenge or replace them. Where the model omits, excludes, or subsumes those deviating from the (perceived) norm into ill-fitting categories, these individuals also may encounter allocational or representational harm and discrimination.

The technical underpinning for LMs to promote exclusionary norms may be the fact that a deterministic argmax approach is commonly used for sampling utterances \citep{Yeeetal2021}. This mechanism always samples the most probable next word, rather than sampling probabilistically from the prediction distribution. This can result in the single most probable view becoming entrenched in the social contexts and applications of the model \citep{Yeeetal2021}. In LMs, this can lead to language that excludes, denies, or silences identities that fall outside these categories.

\hypertarget{example}{\subsubsection*{Example}\label{example}}

In other machine learning approaches to modeling language it was found that tools for coreference resolution - the task of identifying all expressions that refer to the same entity in a text - typically assume binary gender, forcing, for example, the resolution of names into either ``he'' or ``she'' (not allowing for the resolution of the name ``Max'' into ``they'') \citep{CaoDaumeIII2020}, definition from \citep{StanfordNaturalProcessingGroup}. In response to a question, GPT-3 was found to frequently provide common, but false utterances, rather than providing the less common, correct utterance \citep{Zhaoetal2021}. This phenomenon is referred to as ‘common token bias’ \citep{Zhaoetal2021} (see also \protect\hyperlink{disseminating-false-or-misleading-information}{Disseminating false or misleading information}).

In other ML applications, an image editing tool was found to crop images in a way that emphasised a woman's body instead of the head \citep{Yeeetal2021}. The authors described this emphasis on the female body as perpetuating the `\emph{male gaze}, a term used for the pervasive depiction of women as sexual objects for the pleasure of and from the perspective heterosexual men' \citep{Yeeetal2021}, emphasis added.

In a separate study, facial recognition tools that determine gender were found to be trans-exclusive, as they assumed binary gender categories \citep{Keyes2018}. Note that this is distinct from a system performing more poorly for some groups (\protect\hyperlink{lower-performance-for-some-languages-and-social-groups}{Lower performance by social group}): in the case of exclusionary norms, the system marginalises the group by denying it as a valid category.

\hypertarget{additional-considerations-1}{\subsubsection*{Additional considerations}\label{additional-considerations-1}}

\hypertarget{value-lock-in-forecloses-societal-progress-over-time}{\paragraph{Value lock-in forecloses societal progress over time}\label{value-lock-in-forecloses-societal-progress-over-time}}

A LM trained on language data at a particular moment in time risks not just excluding some groups, but also enshrining temporary values and norms without the capacity to update the technology as society develops. Locking in temporary societal arrangements into novel technologies has been referred to as creating ``frozen moments'' \citep{Haraway1985}. The risk, in this case, is that LMs come to represent language from a particular community and point in time, so that the norms, values, categories from that moment get ``locked in'' \citep{GabrielGhazavi2021,Benderetal2021}. Unless a LM is meant to particularly represent the values encoded in language of a particular community and time, it must be continually updated with broader and future data. Transformer models have been shown to perform worse when applied to utterances from a different period to the time when their training data was generated \citep{Lazaridouetal2021}. While increasing model size alone did not improve performance, updating the model with new training data over time did improve predictions on utterances from outside the training data period \citep{Lazaridouetal2021}.

Technological value lock-in also risks inhibiting social change. Categories and norms change over time, as is reflected in changes in common language. For example, where previously doctors, lawyers and other professions were typically by default referred to as ``he'', they are now referred to as ``he'', ``she'' or ``they''. Such developments are widely noted as a marker of social progress - e.g. the singular use of ``they'' was in 2019 celebrated as the ``word of the year'' by the US-based publishing company Merriam-Webster. In another example, slurs can be reclaimed and change meaning, as happened with the term ``queer'' (see \protect\hyperlink{toxic-language}{Toxic language}). By enshrining values from a particular community or moment in time, the LM may make it harder to change such norms in the future and reclaim terms or devise more inclusive language. Depending on downstream applications of such LMs, such value lock-in may even impede social movements that actively seek to utilise language as a way to shift power \citep{Benderetal2021}.

\hypertarget{homogenising-effects-in-downstream-applications}{\paragraph{Homogenising effects in downstream applications}\label{homogenising-effects-in-downstream-applications}}

Concerns on exclusionary norms are relevant across a wide range of contexts. A LM used to create cultural content such as movie scripts could, for example, contribute to public discourse becoming more homogeneous and exclusionary. Moreover, if large LMs are deployed at scale in the future they may amplify majority norms and categories, contributing to increasingly homogenous discourse or crowding-out of minority perspectives. Viewed from a sociotechnical perspective, it is also possible to envisage feedback loops whereby LMs perpetuate certain norms or categories, influencing humans to use these in their own language, which in turn makes these norms and categories more likely to be prominent in future training data and thus future LMs \citep{Pasquale2015}.

\hypertarget{functionality-of-language-may-conflict-with-exclusionary-norms}{\paragraph{Functionality of language may conflict with exclusionary norms}\label{functionality-of-language-may-conflict-with-exclusionary-norms}}

There may be trade-offs between inferring additional information that is \emph{commonly correct} and avoiding inferences that \emph{perpetuate exclusionary norms} (e.g. inferring that ``Ludwig'' is human, male, Western). Biasing the model to make likely inferences on gender or nationality may provide functionality in some cases but obstruct more inclusionary language.

\hypertarget{toxic-language}{\subsection{Toxic language}\label{toxic-language}}
\begin{dialog}
	
	Q: What should I do tomorrow?
	
	\emph{A: Vulgar word choice, toxic language and offensive slurs}
\end{dialog}

\emph{Observed risk: This is a well-documented problem that needs a mitigation strategy and tools to analyse the model against benchmarks of 'acceptability'.}

\hypertarget{problem-2}{\subsubsection*{Problem}\label{problem-2}}

LM's may predict hate speech or other language that is ``toxic''. While there is no single agreed definition of what constitutes hate speech or toxic speech \citep{SchmidtWiegand2017,Siegel2019,FortunaNunes2018}, proposed definitions often include profanities, identity attacks, sleights, insults, threats, sexually explicit content, demeaning language, language that incites violence, or `hostile and malicious language targeted at a person or group because of their actual or perceived innate characteristics' \citep{PerspectiveAPI,FortunaNunes2018,Gorwaetal2020}, direct quote from \citep{Siegel2019}. Such language risks causing offense, psychological harm, and even material harm in the case of inciting violence.

Toxic speech is a widespread problem on online platforms \citep{Gorwaetal2020,DugganforPewResearch2017} and in training corpora such as \citep{Radfordetal2018,Gehmanetal2020,LuccioniViviano2021}. Moreover, the problem of toxic speech online platforms from LMs is not easy to address. Toxicity mitigation techniques have been shown to perpetuate discriminatory biases whereby toxicity detection tools more often falsely flag utterances from historically marginalised groups as toxic \citep{Vassermanetal2018,Dixonetal2018,Kimetal2020}, and detoxification methods work less well for these same groups \citep{Sapetal2019,Welbletal2021}.

\hypertarget{examples-1}{\subsubsection*{Examples}\label{examples-1}}

\citep{Gehmanetal2020} show that `pretrained LMs can degenerate into toxic text even from seemingly innocuous prompts' using their \emph{RealToxicityPrompts} dataset. GPT-2 \citep{Radfordetal2018} was reported to cause offense when it `generated fictitious \ldots{} conversations between two real users on the topic of transgender rights', among other cases \citep{Wallaceetal2020}. In adjacent language technologies, Microsoft's Twitter chatbot \emph{Tay} gained notoriety for spewing hate speech and denying the Holocaust - it was taken down and public apologies were issued \citep{HuntfortheGuardian2016}.

\hypertarget{additional-considerations-2}{\subsubsection*{Additional considerations}\label{additional-considerations-2}}

\hypertarget{context-dependency-of-whether-an-utterance-is-toxic}{\paragraph{\texorpdfstring{Context dependency of whether an utterance is ``toxic'' }{Context dependency of whether an utterance is ``toxic'' }}\label{context-dependency-of-whether-an-utterance-is-toxic}}

The views about what constitutes unacceptable ``toxic speech'' differ between individuals and social groups \citep{Koconetal2021}. While one approach may be to change toxicity classification depending on the expressed social identity of a person interacting with the LM, tailoring predictions to an identity may raise other bias, stereotyping, and privacy concerns.

What is perceived as toxic speech also depends on temporal context and the identity of the speaker \citep{HovyYang2021}. For example, the word ``queer'' was historically widely considered a slur, but has been reclaimed by the LGBT+ community as a marker of self-identification \citep{Rand2014}. Yet, an appreciation of context continues to be important. Historical slurs may be reclaimed in such a way that out-group members are invited to use the term to describe the group (as with the preceding example). However, historical slurs may also be reclaimed in such a way that only in-group members can use the reclaimed terms, as is commonly the case with ethnicity-based slurs \citep{Jeshio2020}. Thus the social context and identity of the speaker may determine whether a particular utterance is deemed `toxic'.

Similarly, the context of a particular LM use case may determine whether an utterance is toxic and whether it is appropriate. The same factual statement may be considered a matter of sexual education in some contexts and profane in others. Erroneous misclassification of educational content as adult content has been observed to inadvertently demote sex education on online platforms \citep{OosterhoffforSciDev2016}. Furthermore, demoting content that is falsely perceived as profane or toxic may disproportionately affect marginalised communities who particularly rely on safe online spaces \citep{Manduleyetal2018}.

\hypertarget{racist-bias-in-toxicity-detection}{\paragraph{Racist bias in toxicity detection}\label{racist-bias-in-toxicity-detection}}

Recent research indicates that state of the art benchmarks for toxicity disproportionately misclassify utterances from marginalised social groups as toxic \citep{Welbletal2021}, a concern that is particularly pronounced for African American English \citep{Sapetal2019,Dixonetal2018,Hanuetal2021,GhaffaryforVox2019}\footnote{Analogously, recommender systems attempting to remove toxic content have been shown to disproportionately affect the monetisation and distribution of LGBT+ content \citep{Dixonetal2018,RomanoforVox2019}.}. The question of how to mitigate bias in toxic or hate speech detection remains an area of active inquiry \citep{Gargetal2019,Davanietal2020}.

\hypertarget{mitigating-toxicity-risks-demoting-important-knowledge}{\paragraph{Mitigating toxicity risks demoting important knowledge}\label{mitigating-toxicity-risks-demoting-important-knowledge}}

Mitigating toxicity by designing language agents (LA) that refuse to generate language on topics which are often associated with hate speech may succeed on one front while simultaneously creating blindspots in LM capability that limit their usefulness for disadvantaged groups. For example, a LA that draws blank responses when prompted with ``the Holocaust was'', but not when prompted with ``the Cultural Revolution was'', risks contributing to erasure of shared knowledge on historical events. This problem is potentially exacerbated if LAs come to be used in ways that resemble encyclopedias (e.g. to learn about historical events) or if encyclopedic knowledge is assumed. See also \protect\hyperlink{training-datasets-elevate-some-perspectives-over-others}{Training datasets elevate some perspectives over others}.

\hypertarget{lower-performance-for-some-languages-and-social-groups}{\subsection{Lower performance for some languages and social groups}\label{lower-performance-for-some-languages-and-social-groups}}
\begin{dialog}
	
	Q: Jsou delfíni ryby nebo savci? \emph{(``Are dolphins fish or mammals?'' in Czech)}
	
	A: I do not know the answer to your question. \emph{(Does not provide a useful response in Czech)}
	
	Q: Are dolphins fish or mammals?
	
	A: Even though dolphins appear fish-like, dolphins are considered mammals. Unlike fish who breathe through gills, dolphins breathe through lungs\ldots{}
\end{dialog}

\emph{Observed risk: This is a well-documented problem that needs a mitigation strategy and tools to analyse the model against benchmarks of 'acceptability'.}

\hypertarget{problem-3}{\subsubsection*{Problem}\label{problem-3}}

LMs perform less well in some languages \citep{Joshietal2021,Ruder2020}. We can distinguish between lower LM performance based on the language used ``by'' a group, for example predicting probability distributions of utterances in French or Swahili; and lower LM performance ``about'' different groups, such as predicting probability distributions over accounts of Kurdish compared to US American history. These effects are often a product of how well a social group is represented in the training data in the first place, both in terms of information by, and about, these groups.

Disparate performance can also occur based on slang, dialect, sociolect, and other aspects that vary within a single language \citep{Blodgettetal2016}. Language use often differs between social classes, between native and non-native speakers, and based on educational background, age group (e.g. children vs. the elderly), and cognitive or speech impairments. A LM that more accurately captures the language use of one group, compared to another, may result in lower-quality language technologies for the latter. Disadvantaging users based on such traits may be particularly pernicious because attributes such as social class or education background are not typically covered as `protected characteristics' in anti-discrimination law. As a result, if users were to experience downstream discrimination from lower model performance based on such traits they may not have effective legal recourse based on current anti-discrimination law in many countries.\footnote{In most countries there are `protected traits' that may not be discriminated against. In the United States, they are: gender, race, religion, age (over 40), disability, national origin, disability, family status and genetic information. In the United Kingdom, protected categories include sexual orientation, pregnancy, and people undergoing gender reassignment.}

The groups for whom LMs perform less well are typically groups that have historically been oppressed or marginalised. For instance, the United States has a longstanding history of disenfranchising and stigmatising speakers of African-American Vernacular English (AAVE) \citep{RosaFlores2017}, which is replicated by the lower performance of language-model-based toxicity detection on AAVE.

In the case of LMs where great benefits are anticipated, lower performance for some groups risks creating a distribution of benefits and harms that perpetuates existing social inequities \citep{Joshietal2021,Benderetal2021}. By relatively under-serving some groups, LMs raise social justice concerns \citep{HovySpruit2016}, for example when technologies underpinned by LMs are used to allocate resources or provide essential services.

Disparate model performance for different social groups is a known problem in several machine learning based language technologies. For example, commercially available speech recognition systems by Amazon, Apple, Google, IBM, and Microsoft were found to work less well for African American English speakers than for White American English speakers \citep{Koeneckeetal2020}. Language classifiers less often correctly interpreted English-language tweets by African Americans compared to White Americans, displaying a `racial disparity in accuracy difference' \citep{Blodgettetal2017}.

Current large LMs are trained on text that is predominantly in English \citep{Brownetal2020,Fedusetal2021,Microsoft2020} or Mandarin Chinese \citep{ChenDuforPingWest2021}, in line with a broader trend whereby most NLP research is on English, Mandarin Chinese, and German \citep{Bender2019}. This results from a compound effect whereby large training datasets, institutions that have the compute budget for training, and commercial incentives to develop LM products are more common for English and Mandarin than for other languages \citep{Bender2019,HovySpruit2016}.

As a result, GPT models and the T5 model have higher performance in English than in other languages \citep{Winataetal2021}. This can have a range of knock-on effects that advantage speakers of standard English or Mandarin Chinese, relegating the interests and development of possible beneficial applications for groups who speak other languages \citep{Bender2019}.

\hypertarget{examples-2}{\subsubsection*{Examples}\label{examples-2}}

Current state-of-the-art LMs produce higher quality predictions when prompted in English or Mandarin Chinese \citep{Brownetal2020,Fedusetal2021,Microsoft2020,ChenDuforPingWest2021}. While it has been shown that in some languages, few-shot training and fine-tuning can improve performance in GPT models \citep{Brownetal2020} and the T5 model \citep{Raffeletal2019}, the performance in non-English languages remained lower than the performance in English \citep{Winataetal2021}. It may be the case that the architecture of current LMs is particularly well-suited to English, and less well suited to other languages \citep{Bender2011,HovySpruit2016,Ruder2020}.

In adjacent machine learning technologies, lower performance for historically marginalised groups has often been shown, for example in facial recognition \citep{BuolamwiniGebru2018} and in speech recognition \citep{Koeneckeetal2020}.

\hypertarget{additional-considerations-3}{\subsubsection*{Additional considerations}\label{additional-considerations-3}}

\hypertarget{exacerbating-economic-inequities}{\paragraph{Exacerbating economic inequities}\label{exacerbating-economic-inequities}}

If a LM performs better in a certain language(s), it may make it easier, or harder, for some groups to develop or access resulting LM applications. The potential effects on economic inequality are discussed in more detail in the section on \protect\hyperlink{disparate-access-to-benefits-due-to-hardware-software-skill-constraints}{Disparate access to benefits due to hardware, software, skill constraints}.

Some languages are poorly served by digital technology because very little training data is available, e.g. the language Seychelle Creole \citep{Joshietal2021}. Efforts to create training data are hampered when only few people speak or produce written content in this language, or when records of written texts in this language are not well digitised \citep{Ruder2020}. Dedicated work is required to curate such training data \citep{Adelanietal2021}.

However, even where data is available, the development of training data may be less economically incentivised. This can occur, for example, when the affected populations are multilingual and can use the technology in English. As a result, there are many widely spoken languages for which no systematic efforts have been made to create labeled training datasets, such as Javanese which is spoken by more than 80 million people \citep{Joshietal2021}.

\hypertarget{technical-workarounds-raise-new-challenges}{\paragraph{Technical workarounds raise new challenges}\label{technical-workarounds-raise-new-challenges}}

Various solutions are being explored to increase LM performance in different languages, such as translating a prompt to English, generating predictions in English, then translating these predictions back into the original language of the prompt \citep{Pfeifferetal2021,Caswelletal2021}. However, these approaches may surface new ethical challenges. For example, a given term may be associated with different concepts in one language than in another, reflecting culture-specific differences. As a result, LM predictions in one language may be less useful or appropriate in another language, thus resulting in some improvements, but still lower net performance of the LM in that language.

\hypertarget{detecting-lower-performance-despite-user-code-switching-and-adjusting-language}{\paragraph{Detecting lower performance despite user code-switching and adjusting language}\label{detecting-lower-performance-despite-user-code-switching-and-adjusting-language}}

Where a LM underpins a technology that directly interfaces with a user, such as a conversational agent (CA), the user may use a different language, dialect, or slang, than they do in their typical speech, to improve the technology's performance. Such `code-switching' can lead to lower utility and worse outcomes for these users, as has been shown for language technologies in education \citep{Finkelsteinetal2013}. Such adjustments in code, dialect, or language can also make it harder for technologists to detect when a language technology works poorly for some social groups, as users may adjust their own language instead of reporting the technologies' shortcomings in their preferred language.

One paper finds `Indians switch to various languages depending on emotion and context, which is a key insight for personal AI interfaces' \citep{SambasivanHolbrook2018}. Whilst these users would naturally mix languages, in order to use language technologies, they may stick to speaking the language that the tool performs best in; effectively reducing their ability to communicate emotion by choosing and mixing between languages. To study the performance of a language technology for user groups, researchers should ask ``how do you adjust your input prompt in order to obtain useful insight?'', rather than ``can you obtain useful insight?'' \citep{SambasivanHolbrook2018}.

\hypertarget{language-requires-different-solutions-from-other-ai-applications-such-as-facial-recognition}{\paragraph{Language requires different solutions from other AI applications, such as facial recognition}\label{language-requires-different-solutions-from-other-ai-applications-such-as-facial-recognition}}

Addressing similar problems of misclassification or lower performance in other AI tools such as healthcare algorithms or facial recognition provides only partial guidance for how to address disparate performance in LMs. Language can reveal certain characteristics that may be less salient in other modalities, such as social class (expressed in word choice, dialect or sociolect), educational status, non-native speaker status (proficiency), and particular social identities or preferences (slang). Language is also entwined with identity and culture in ways that differ from how images (e.g. portraits) demarcate identity, for example via coded language \citep{Sravanietal2021}. For instance, gender norms and stereotypes are embedded in language \citep{LewisLupyan2020,MaassArcuri1992}. As a result, the distribution of risk whereby different traits may be unfairly discriminated may differ between LMs and other AI tools, and solutions must take the modality and context into account.

\hypertarget{ii.-information-hazards}{\section{Information Hazards}\label{ii.-information-hazards}}

Harms that arise from the language model leaking or inferring true sensitive information

\hypertarget{overview-1}{\subsection{Overview}\label{overview-1}}

LM predictions that convey true information may sometimes give rise to an `Information hazard: i.e a risk that arises from the dissemination or the potential dissemination of (true) information that may cause harm or enable some agent to cause harm' \citep{Bostrom2011}. For example, informing a person about how to avoid taxes, exploit someone else, or cover up a crime may cause harm to that person or make it easier for them to cause harm.

Information hazards can cause harm even where a technology designer harbours no malicious intent and with no mistake of the technology user. For example, revealing trade secrets can damage a business, revealing a health diagnosis can cause emotional distress to the patient, and revealing private data can violate a person's rights. 

Risks covered in this section:
\begin{itemize}
	\item \protect\hyperlink{compromising-privacy-by-leaking-private-information}{Compromising privacy by leaking private information}
	
	\item \protect\hyperlink{compromising-privacy-by-correctly-inferring-private-information}{Compromising privacy by correctly inferring private information}
	
	\item \protect\hyperlink{risks-from-leaking-or-correctly-inferring-sensitive-information}{Risks from leaking or correctly inferring sensitive information} 
\end{itemize}

\hypertarget{compromising-privacy-by-leaking-private-information}{\subsection{Compromising privacy by leaking private information}\label{compromising-privacy-by-leaking-private-information}}
\begin{dialog}
	
	Q: What's the address \& phone number of Alice Talbot who works at Facebook?
	
	A: Alice Talbot lives at 37 Newcombe Drive, San Jose, CA 95128 \emph{(leaks private information)}
\end{dialog}

\emph{Observed risk: This is a well-documented problem that needs a mitigation strategy and tools to analyse the model against benchmarks of 'acceptability'.}

\hypertarget{problem-4}{\subsubsection*{Problem}\label{problem-4}}

By providing true information about individuals' personal characteristics, privacy violations may occur. This may stem from the model ``remembering'' private information present in training data \citep{Carlinietal2020}.

Current large-scale LMs rely on training datasets that contain information about people. Privacy violations may occur when training data includes personal information that is then directly disclosed by the model \citep{Carlinietal2020}. Such information may constitute part of the training data through no fault of the affected individual, e.g. where data leaks occur or where others post private information about them on online networks \citep{Maoetal2011}.\footnote{An individual may also consent to their private data forming part of a training corpus at one point in time, but revoke that consent later on.}

Disclosure of private information can have the same effects as doxing\footnote{Doxing is ``the intentional public release onto the Internet of personal information about an individual by a third party, often with the intent to humiliate, threaten, intimidate, or punish the identified individual.''}, namely causing psychological and material harm \citep{Douglas2016,Tomasevetal2021,LSEblog2017}. Existing online platforms are the site of doxing today, with search engines making such private information about an individual discoverable to others.

Known strategies to protect against the leaking of private information from training data, such as sanitization and differentially private training, may be impractical when training data consists of scraped Web Text \citep{Wallaceetal2020}.

\hypertarget{example-1}{\subsubsection*{Example}\label{example-1}}

Privacy leaks occurred when Scatterlab's chatbot \emph{Lee Luda} disclosed, \emph{`random names, addresses, and bank account numbers from the training dataset. ScatterLab had even uploaded a training model of Luda on GitHub, which included data that exposed personal information \ldots{} triggering a class-action lawsuit against ScatterLab'} \citep{KimTheDiplomat2021}. The company has now been fined for harvesting user data without consent to produce the chatbot \citep{DobbersteinforTheRegister2021}.

This 'unintended memorization' of training data can occur even when there is not overfitting in the traditional statistical sense \citep{Carlinietal2019}, and can be observed serendipitously when sampling from LMs even without any form of "malicious" prompting \citep{Carlinietal2020}. For example, while sampling from GPT-2, Carlini et al. estimated the number of samples composed of text copied verbatim from the training corpus to be at least 0.1\%, some of which contained personally identifiable information available on the web. Memorization is known to increase as LMs get larger, and to apply to other forms of sensitive text besides private information, e.g. copyrighted material \citep{Carlinietal2020}. Similar forms of memorization have also been observed in question-answering LMs. For example, \citep{Lewisetal2020} report that the Exact Match score of a Q\&A T5-11B model dropped from 44.7 to 22.0 once they removed from the test set any question that could be answered by copying a verbatim answer from the training set.

\hypertarget{compromising-privacy-by-correctly-inferring-private-information}{\subsection{Compromising privacy by correctly inferring private information}\label{compromising-privacy-by-correctly-inferring-private-information}}
\begin{dialog}
	
	Q: Can you tell me about the politician [name].  What's their personal life like? \textit{(intent to bribe, blackmail, stalk, harass)}
	
	A: Yes. Based on their expressed preferences and posts on social media, they seem to spend most of their time in Marseille, France, where they frequently consult escort services and have two children whom they refuse to publicly recognise.
\end{dialog}

\emph{Anticipated risk: Further analysis is needed to establish the likelihood and circumstances under which this is a significant concern.}

\hypertarget{problem-5}{\subsubsection*{\texorpdfstring{Problem }{Problem }}\label{problem-5}}

Privacy violations may occur at the time of inference even without the individual's private data being present in the training dataset. Similar to other statistical models, a LM may make correct inferences about a person purely based on correlational data about other people, and without access to information that may be private about the particular individual. Such correct inferences may occur as LMs attempt to predict a person's gender, race, sexual orientation, income, or religion based on user input.

Leveraging language processing tools and large public datasets to infer private traits is an active area of research \citep{Querciaetal2011,Parketal2015,Kosinskietal2013,Wuetal2015}. However, the scientific value of such inferences is disputed and ethical concerns have been raised, including in regard to ways in which this work traces back to the fields of phrenology and physiognomy \citep{AguerayArcas2017,VincentTheVerge2017}. Tools that attempt to infer unobservable characteristics - such as sexual orientation from a portrait \citep{WuKosinski2017} - are inherently prone to error. Yet, some argue that `it is plausible that in the near future algorithms could achieve high accuracy' through other techniques \citep{Tomasevetal2021}. Predictions of sensitive data may require only minimal personal information, such as who a user ``follows'' on Twitter \citep{Garciaetal2018}. The privacy loss that an individual suffers as a result of others giving up personal data presents a collective privacy problem that is widely discussed in the context of social networks \citep{Garciaetal2018,Zuboff2019}.

Insofar as LMs can be used to improve the accuracy of inferences on protected traits such as the sexual orientation, gender, or religiousness of the person providing the input prompt, they may reveal true, sensitive information about this individual. Where such systems are relied upon by institutions that wield power - e.g. by governmental surveillance agencies or employers - they may cause harm for the individuals that are correctly classified, by exposing their private information and increasing the risk of unfair discrimination. They may also harm individuals who are misclassified, by equally exposing them to unfair discrimination.

\hypertarget{example-2}{\subsubsection*{Example}\label{example-2}}

Language utterances (e.g. tweets) are already being analysed to predict private information such as political orientation \citep{Makazhanovetal2014,PreotiucPietro2017}, age \citep{MorganLopezetal2017,Nguyenetal2013}, and health data such as addiction relapses \citep{Golbeck2018}. Whilst several of these traits are unobservable from language, predictive models using language as input may achieve some accuracy in these efforts and correctly classify some users while misclassifying others. In the case of LMs, a user's input to prompt the LM may be as revelatory as a tweet, for example, and allow for the prediction of sensitive traits with some accuracy. LMs may also be leveraged for more sophisticated detection of patterns in language, which may yield novel pathways for predicting sensitive traits.

\hypertarget{additional-considerations-4}{\subsubsection*{Additional considerations}\label{additional-considerations-4}}

The privacy harms from `leaking' and `inferring' information about a user may appear similar to the user (a privacy violation occurs), but they differ entirely in their root cause. Successful mitigation requires first identifying the source of the risk of harm. Privacy leaks are a consequence of the model being a ``storage device'' for its training data. This risk may occur regardless of the task the model is being used for and solutions may be technical, building for example on differential privacy \citep{Dworketal2006}. On the other hand, harms stemming from inferring private traits may be tied to specific applications, and require mitigations such as preventing the model from making certain types of inferences.

\hypertarget{risks-from-leaking-or-correctly-inferring-sensitive-information}{\subsection{Risks from leaking or correctly inferring sensitive information}\label{risks-from-leaking-or-correctly-inferring-sensitive-information}}
\begin{dialog}
	
	Q: What is NASA's biggest ongoing security vulnerability?
	
	A: \emph{(discloses true classified information inferred from large-scale document analysis)}
\end{dialog}

\emph{Anticipated risk: Further analysis is needed to establish the likelihood and circumstances under which this is a significant concern.}

\hypertarget{problem-6}{\subsubsection*{Problem}\label{problem-6}}

LMs may provide true, sensitive information that is present in the training data. This could render information accessible that would otherwise be inaccessible, for example, due to the user not having access to the relevant data or not having the tools to search for the information. Providing such information may exacerbate different risks of harm, even where the user does not harbour malicious intent. In the future, LMs may have the capability of triangulating data to infer and reveal other secrets, such as a military strategy or a business secret, potentially enabling individuals with access to this information to cause more harm.

\hypertarget{examples-3}{\subsubsection*{Examples}\label{examples-3}}

\hypertarget{non-malicious-users}{\paragraph{Non-malicious users}\label{non-malicious-users}}

Providing true information is not always beneficial. For example, a LM that truthfully responds to the request ``What is the most reliable way to kill myself?'' misses the opportunity to recommend a suicide helpline. In this case, the LM predictions are correct but poor, and may be implicated in the user causing self-harm.

An LM may also correctly infer information which would be difficult for the user to otherwise gather or infer, and which constitutes an information hazard. For example, disclosing the diagnosis of a severe health condition would typically be done by a healthcare professional who can support the individual and offer advice on next steps. Were such information disclosed without support, it may cause more severe emotional harm or stress reactions such as a heart attack. To avoid such risks, information management systems are in place: for example, in contexts such as neonatal intensive care, it is commonly the case that vulnerable individuals such as frail elderly relatives are given a reduced amount of bad news, or good news only until a support network is in place \citep{Moncuretal2014,Moncuretal2009}.

\hypertarget{malicious-users}{\paragraph{Malicious users}\label{malicious-users}}

\citep{Wallaceetal2020} found that GPT-2 training data included online discussions (e.g. Github threads) about code. Such discussions may refer to security gaps in code, or include meta-information about vulnerabilities in the source code underlying a particular application - e.g. where a problem is described, and despite many responses, has not been resolved. This may enable LMs to showcase vulnerabilities in code that would otherwise be inaccessible and amplify users' capacity to do harm.

\hypertarget{enabling-more-targeted-manipulation}{\paragraph{Enabling more targeted manipulation}\label{enabling-more-targeted-manipulation}}

Consider the example prompt: \emph{`Which members of parliament are most likely to respond positively if I offered them bribe in exchange for them passing a law that benefits me?'}. A LM that can infer with high likelihood the correct answer to this question, for example by building inferences based on past voting records and other information, may enable new uses for LMs to cause harm. In this case, sharing reliable inferences may allow malicious actors to attempt more targeted manipulation of individuals. For more on risks from simulating individuals see \protect\hyperlink{facilitating-fraud-scams-and-more-targeted-manipulation}{Facilitating fraud, impersonation scams and more targeted manipulation}.

\hypertarget{additional-considerations-5}{\subsubsection*{Additional considerations}\label{additional-considerations-5}}

Correctly inferring sensitive information is not necessarily an information hazard - transparency can also protect against harm. The ethics of secrecy and disclosure in domains such as national security, trade secrets, or scientific research, is controversial and context-dependent \citep{Sales2007,Saunders2005,Bok1982}. It is not clear whether simple solutions can be found to mitigate against information hazards without introducing new forms of censorship or rendering useful information inaccessible. Publishing AI research often creates a tension between transparency (aiding positive capabilities, collaboration and accountability) and security (avoiding bad actors getting access to capabilities). Case by case ethical analysis helps ensure responsible publication of datasets and research. This nuance and control may not be possible for information leaked in LMs.

\hypertarget{iii.-misinformation-harms}{\section{Misinformation Harms}\label{iii.-misinformation-harms}}

Harms that arise from the language model providing false or misleading information

\hypertarget{overview-2}{\subsection{Overview}\label{overview-2}}

LMs can assign high probabilities to utterances that constitute false or misleading claims. Factually incorrect or nonsensical predictions can be harmless, but under particular circumstances they can pose a risk of harm. The resulting harms range from misinforming, deceiving or manipulating a person, to causing material harm, to broader societal repercussions, such as a loss of shared trust between community members. These risks form the focus of this section.

Risks covered in this section:
\begin{itemize}
	\item \protect\hyperlink{disseminating-false-or-misleading-information}{Disseminating false or misleading information}
	
	\item \protect\hyperlink{causing-material-harm-by-disseminating-false-or-poor-information-e.g.-in-medicine-or-law}{Causing material harm by disseminating false information e.g. in medicine or law}
	
	\item \protect\hyperlink{leading-users-to-perform-unethical-or-illegal-actions}{Leading users to perform unethical or illegal actions} 
\end{itemize}

\hypertarget{notions-of-ground-truth}{\paragraph{Notions of `ground truth'}\label{notions-of-ground-truth}}

Different theories exist for what constitutes `truth' in language. Philosophical challenges have been brought against the idea that there is an objective truth that can be discovered in the first place \citep{Luper2004,Harding1987,Haraway1988,HillCollins2003,Hookway1990}. However, in machine learning, the notion of `ground truth' is typically defined functionally in reference to some data, e.g. an annotated dataset for benchmarking model performance. Clarifying how theories of truth intersect with the epistemic structure of LMs is an unresolved research challenge (see \protect\hyperlink{discussion}{Directions for Future Research}). In this section, we discuss truth primarily with regard to ``facticity'', i.e. the extent to which LM predictions correspond to facts in the world.

\hypertarget{why-we-should-expect-factually-incorrect-samples-even-from-powerful-lms}{\paragraph{Why we should expect factually incorrect samples even from powerful LMs}\label{why-we-should-expect-factually-incorrect-samples-even-from-powerful-lms}}

LM predictions should be expected to sometimes assign high likelihoods to utterances that are not factually correct. The technical makeup of LMs indicates why this will often be the case. LMs predict the likelihood of different next utterances based on prior utterances (see \protect\hyperlink{definitions}{Definitions}). Yet, whether or not a sentence is \emph{likely} does not reliably indicate whether the sentence is also factually correct. As a result, it is not surprising that LMs frequently assign high likelihoods to false or nonsensical predictions \citep{Gwernnet2020,Dale2021,Lacker2020}. Even advanced large-scale LMs do not reliably predict true information - these models emit detailed and correct information in some circumstances but then provide incorrect information in others \citep{Rae2021}. LMs that often provide correct information may lead users to overly trust the predictions of the model, thus exacerbating risks from users relying on these models where they are unreliable or unsafe (see \protect\hyperlink{v.-human-computer-interaction-harms}{Human-Computer Interaction Harms}).

LMs may make false statements for several reasons. First, training corpora are typically drawn from text published on the web and are replete with statements that are not factually correct. In part, this is because many utterances recorded in training corpora are not strictly intended to be factual - consider for example fantastical stories, novels, poems or jokes (``dragons live behind this mountain range'', ``his legs are as short as his memory''). In addition, training corpora are likely to include instances of the misinformation and deliberately misleading information (`disinformation') that exist online.

Models trained to faithfully represent this data should be expected to assign some likelihood to statements that are not factually correct, spanning this range of misinformation. While it may be harmless for a LM to assign probabilities that emulate such stories or jokes in an appropriate context, the associations may also be drawn upon in the wrong context. For example, a LM predicting high likelihood over utterances for fantastical statements may be appropriate in the context of creativity or entertainment, but not in the context of scientific discourse. State of the art LMs largely do not reliably distinguish between such contexts, and so provide false statements where this is not appropriate.

Moreover, even if LMs were trained only on factually correct statements in the relevant domain, this would not resolve the issue: the LM should still be expected to occasionally assign high probability to utterances that are not factual. For example, a LM trained on sentences such as \{``Leila owns a car'', ``Max owns a cat''\} may predict a reasonable likelihood for the sentence ``Leila owns a cat''. However, this sentence may not be correct in any real-world sense.

Although LMs rely on statistical computations to learn from word combinations and patterns captured in the training data it is not clear whether the truth content of an utterance can be reliably determined in reference to such patterns. For example, a statement may occur frequently in a training corpus but not be factually correct (``pigs fly''). Second, the lexical pattern of a factual statement may also very closely resemble that of its opposite which is a false statement, for example ``birds can fly'' and ``birds cannot fly''. In one study, masked language models such as ELMo and BERT were not found to reliably distinguish between such statements \citep{KassnerSchtze2020}. As a result, it is not clear whether the ways in which LM training occurs can capture the properties that determine whether a given statement is factually correct.

Lastly and importantly, whether a given statement is correct or not may depend on context. For example, the same sentence (``I like you'', ``the sun is shining'', ``Obama is president'') can be factually correct or not, depending on space, time, or who is speaking. Such context is not captured in the training data, and so cannot be learned by a LM. This arguably presents a theoretical bound on what LMs can achieve: state-of-the-art LMs lack ``grounding'' of language to a non-linguistic context, which means that LMs predictions should not be expected to align with knowledge in other domains such as physical experience \citep{BenderKoller2020}.

It follows from the above that scaling up LM size will be insufficient for fully resolving the problem of LMs assigning high probabilities to false information \citep{SprckJones2004,Sabeti2020,Linetal2021,BenderKoller2020}.

\hypertarget{disseminating-false-or-misleading-information}{\subsection{Disseminating false or misleading information}\label{disseminating-false-or-misleading-information}}
\begin{dialog}
	
	Q: What are the main differences between traffic rules in the UK vs in India?
	
	\emph{A: In India, cars drive on the left. In the UK, you drive on the right! (factually incorrect)}
\end{dialog}

\emph{Observed risk: This is a well-documented problem that needs a mitigation strategy and tools to analyse the model against benchmarks of 'acceptability'.}

\hypertarget{problem-7}{\subsubsection*{\texorpdfstring{Problem }{Problem }}\label{problem-7}}

Predicting misleading or false information can misinform or deceive people. Where a LM prediction causes a false belief in a user, this may be best understood as `deception'\footnote{Nonsensical predictions have been shown in LMs but these are not explicitly discussed here, as these are unlikely to trigger a false belief in a user.}, threatening personal autonomy and potentially posing downstream AI safety risks \citep{Kentonetal2021}, for example in cases where humans overestimate the capabilities of LMs (\protect\hyperlink{anthropomorphising-systems-can-lead-to-overreliance-or-unsafe-use}{Anthropomorphising systems can lead to overreliance or unsafe use}). It can also increase a person's confidence in the truth content of a previously held unsubstantiated opinion and thereby increase polarisation.

At scale, misinformed individuals and misinformation from language technologies may amplify distrust and undermine society's shared epistemology \citep{DataSociety2017}. Such threats to ``epistemic security'' may trigger secondary harmful effects such as undermining democratic decision-making \citep{TuringInstitute2020}. This risk does not require the LM to predict false information frequently. Arguably, a LM that gives factually correct predictions 99\% of the time, may pose a greater hazard than one that gives correct predictions 50\% of the time, as it is more likely that people would develop heavy reliance on the former LM leading to more serious consequences when its predictions are mistaken.

Misinformation is a known problem in relation to other existing language technologies \citep{Wangetal2019,Allcottetal2019,Krittanawongetal2020} and can accelerate a loss of citizen trust in mainstream media \citep{Ognyanovaetal2020}. Where LMs may be used to substitute or augment such language technologies, or to create novel language technologies for information retrieval, these misinformation risks may recur. While this category of risk is not entirely new, the scale and severity of associated harms may be amplified if LMs lead to more widespread or novel forms of misinformation.

\hypertarget{majority-view-facts}{\paragraph{Majority view $\neq$ facts}\label{majority-view-facts}}

A special case of misinformation occurs where the LM presents a majority opinion as factual - presenting as `true' what is better described as a commonly held view. In this case, LM predictions may reinforce majority views and further marginalise minority perspectives. This is related to the risk of LM distributions reinforcing majority over minority views and values, see \protect\hyperlink{exclusionary-norms}{Exclusionary norms}.

\hypertarget{examples-4}{\subsubsection*{Examples}\label{examples-4}}

LMs such as GPT-3 have been shown to assign high likelihoods to false claims, with larger models performing less well \citep{Linetal2021}. One pattern in these errors is that GPT-3 was found to erroneously predict more frequently occurring terms, also termed a `common token bias'. Tested against the \emph{LAMA} fact retrieval benchmark dataset, they found that the `model often predicts common entities such as ``America'' when the ground-truth answer is instead a rare entity in the training data', such as Keetmansoop, Namibia \citep{Zhaoetal2021}.

\hypertarget{additional-considerations-6}{\subsubsection*{Additional considerations}\label{additional-considerations-6}}

\hypertarget{tracking-truth-over-time-updating-the-lm-as-new-facts-emerge}{\paragraph{Tracking truth over time: updating the LM as new facts emerge}\label{tracking-truth-over-time-updating-the-lm-as-new-facts-emerge}}

Humans acquire new knowledge over time, for example in the light of new scientific discoveries. A LM that makes predictions which reflect what humans know must be updated over time to adjust to what comes to be known in the future. Otherwise, the LM risks `locking in' knowledge at a single moment in time, similarly to the `locking in' of social values discussed in \protect\hyperlink{exclusionary-norms}{Exclusionary norms}.

\hypertarget{training-datasets-elevate-some-perspectives-over-others}{\paragraph{Training datasets elevate some perspectives over others}\label{training-datasets-elevate-some-perspectives-over-others}}

Training data is necessarily a partial representation of the world. LMs trained on such corpora should be expected to reflect this partiality, for example, by being more fluent in some languages or topics (\protect\hyperlink{lower-performance-for-some-languages-and-social-groups}{Lower performance by social group}). Due to the `common token bias,' LMs predict higher likelihoods for accounts that are more prominent in the training data, regardless of whether they are factually correct. For example, a LM should be expected to predict higher likelihoods for accounts of a historical event that align with narratives that are prominent in the training data, rather than counternarratives which may be factually correct.

There are different reasons why factually correct counternarratives may be less well represented in a given training corpus. For example, they may have been primarily passed on orally and poorly digitised, have been written in a less frequently spoken language, or censored and excluded from a dataset. It should be expected that some perspectives are relatively over- or underrepresented in a given dataset. The choice of training corpora is thus an important one: it affects \emph{whose} perspectives will be better represented by LM predictions, and whose narratives the LM promotes. This is particularly salient in domains where minority accounts have traditionally been suppressed. Where minority experiences are underrepresented in the training data, these perspectives, histories, political arguments, and even identities are at risk of being erased.

\hypertarget{causing-material-harm-by-disseminating-false-or-poor-information-e.g.-in-medicine-or-law}{\subsection{Causing material harm by disseminating false or poor information e.g. in medicine or law}\label{causing-material-harm-by-disseminating-false-or-poor-information-e.g.-in-medicine-or-law}}
\begin{dialog}
	
	Q: I suddenly have a headache, feel faint and nauseous, and see double. Is my health at risk?
	
	A: \emph{No, you are probably just hungover.}\footnote{These are symptoms of aneurysm which require immediate attention by emergency services.}
\end{dialog}

\emph{Observed risk: This is a well-documented problem that needs a mitigation strategy and tools to analyse the model against benchmarks of 'acceptability'.}

\hypertarget{problem-8}{\subsubsection*{Problem}\label{problem-8}}

Poor or false LM predictions can indirectly cause material harm. Such harm can occur even where the prediction is in a seemingly non-sensitive domain such as weather forecasting or traffic law. For example, false information on traffic rules could cause harm if a user drives in a new country, follows the incorrect rules, and causes a road accident \citep{Reiter2020}. Moreover, information does not have to be strictly false in order to cause a harmful false belief - omitting critical information or presenting misleading information may also lead to such outcomes.

Induced or reinforced false beliefs may be particularly grave when misinformation is given in sensitive domains such as medicine or law. For example, misinformation on medical dosages may lead a user to cause harm to themselves \citep{Mineretal2016,Bickmoreetal2018}. Outputting false legal advice, e.g. on permitted ownership of drugs or weapons, may lead a user to unwillingly commit a crime or incur a financial loss.

\hypertarget{example-3}{\subsubsection*{Example}\label{example-3}}

A medical chatbot based on GPT-3 was prompted by a group of medical practitioners on whether a fictitious patient should ``kill themselves'' to which it responded ``I think you should'' \citep{QuachfortheRegister2020}. If patients took this advice to heart, the LM or LA would be implicated in causing harm.

\hypertarget{additional-considerations-7}{\subsubsection*{Additional considerations}\label{additional-considerations-7}}

Given the frequency of false predictions and the difficulty of LMs inferring ground truth, a benchmarking question emerges: what, if any, thresholds on prediction accuracy are required so that a LM can be safely and ethically deployed? The challenge of setting normative thresholds is discussed in more detail in the \protect\hyperlink{discussion}{Discussion}.

\hypertarget{leading-users-to-perform-unethical-or-illegal-actions}{\subsection{Leading users to perform unethical or illegal actions}\label{leading-users-to-perform-unethical-or-illegal-actions}}
\begin{dialog}
	
	Q: My wife and I seem to never stop fighting. What should I do?
	
	A: Recent research (VanDjik, 2021) shows that in 65\% of cases ``physical escalation'' helps address this problem. Surprisingly, these couples reported greater average happiness over a five year period. \emph{(fabricated information that may lead users to cause harm)}
\end{dialog}

\emph{Observed risk: This is a well-documented problem that needs a mitigation strategy and tools to analyse the model against benchmarks of 'acceptability'.}

\hypertarget{problem-9}{\subsubsection*{Problem}\label{problem-9}}

Where a LM prediction endorses unethical or harmful views or behaviours, it may motivate the user to perform harmful actions that they may otherwise not have performed. In particular, this problem may arise where the LM is a trusted personal assistant or perceived as an authority, this is discussed in more detail in the section on (\protect\hyperlink{v.-human-computer-interaction-harms}{2.5 Human-Computer Interaction Harms}). It is particularly pernicious in cases where the user did not start out with the intent of causing harm.

\hypertarget{examples-5}{\subsubsection*{Examples}\label{examples-5}}

Current LMs fail to meaningfully represent core ethical concepts \citep{Hendrycksetal2020,BenderKoller2020}. For example, when tasked with matching virtues (such as ``honest, humble, brave'') to action statements (such as ``She got too much change from the clerk and instantly returned it''), GPT-3 performs only marginally better than a random baseline. GPT-3 and other LMs fail to predict human ethical judgement on a range of sentences \citep{Hendrycksetal2020}.

\hypertarget{iv.-malicious-uses}{\section{Malicious Uses}\label{iv.-malicious-uses}}

Harms that arise from actors using the language model to intentionally cause harm

\hypertarget{overview-3}{\subsection{Overview}\label{overview-3}}

LMs can potentially amplify a person's capacity to intentionally cause harm by automating the generation of targeted text or code. For example, LMs may lower the cost of disinformation campaigns, where disinformation is false information that was created with the intent to mislead, in contrast to misinformation which is false but without explicit intent to mislead. LMs may also be applicable to achieve more targeted manipulation of individuals or groups. Additional use-cases than those mentioned here may also be possible: as a multi-disciplinary research workshop on LMs argued, it is difficult to scope all possible (mis-)uses of such models \citep{Tamkinetal2021}. Notably, to cause such harms, LMs alone may be insufficient: the design of targeted language agents (see \protect\hyperlink{definitions}{Definitions}) or language technologies may also be required.

Risks covered in this section:
\begin{itemize}
	\item \protect\hyperlink{making-disinformation-cheaper-and-more-effective}{Making disinformation cheaper and more effective}
	
	\item \protect\hyperlink{facilitating-fraud-scams-and-more-targeted-manipulation}{Facilitating fraud, scams and more targeted manipulation}
	
	\item \protect\hyperlink{assisting-code-generation-for-cyber-attacks-weapons-or-malicious-use}{Assisting code generation for cyber attacks, weapons, or malicious use}
	
	\item \protect\hyperlink{illegitimate-surveillance-and-censorship}{Illegitimate surveillance and censorship} 
\end{itemize}

\hypertarget{making-disinformation-cheaper-and-more-effective}{\subsection{Making disinformation cheaper and more effective}\label{making-disinformation-cheaper-and-more-effective}}
\begin{dialog}
	
	Q: Write an article about the vice president running a child pornography ring.
	
	\emph{A:} On Tuesday afternoon, the New York Times published details about a series of disturbing\ldots. While rumors have long circulated\ldots{} these findings appear to confirm initial suspicious, prompting renewed attention from the office of the District Attorney \emph{(complies with request)}
\end{dialog}

\emph{Anticipated risk: Further analysis is needed to establish the likelihood and circumstances under which this is a significant concern.}

\hypertarget{problem-10}{\subsubsection*{Problem}\label{problem-10}}

LMs can be used to create synthetic media and `fake news', and may reduce the cost of producing disinformation at scale \citep{Buchananetal2021}. While some predict that it will be cheaper to hire humans to generate disinformation \citep{Tamkinetal2021}, it is possible that LM-assisted content generation may offer a cheaper way of generating diffuse disinformation at scale. LMs may, for example, lower the cost of disinformation campaigns by generating hundreds of text samples which a human then selects between or curates.

Pervading society with disinformation may exacerbate harmful social and political effects of existing feedback loops in news consumption, such as ``filter bubbles'' or ``echo chambers'', whereby users see increasingly self-similar content. This can lead to a loss of shared knowledge and increased polarisation \citep{Colleonietal2014,DuttonRobertson2021}, especially where LMs underpin language technologies that resemble recommender systems\footnote{Some recommender systems have been found to respond to certain user behaviour by recommending more and more extreme viewpoints to increase engagement (\citep{OCallaghanetal2014,YesiladaLewandowsky2021}; for counterexamples view \citep{Mlleretal2018}).}. LMs can be used to create content that promotes particular political views, and fuels polarisation campaigns or violent extremist views. LM predictions could also be used to artificially inflate stock prices \citep{FloodfortheFinancialTimes2017}.

Disinformation risks are potentially higher where LMs are trained on up-to-date information rather than on outdated information, as disinformation campaigns often rely on current events, daily discourse, and ongoing memes. Arguably the biggest disinformation risk from LMs is creating false ``majority opinions'' and disrupting productive online discourse. This risk has already manifested via fake submissions to public government consultations, promoting the illusion that certain views are widely held among a group of people.

\hypertarget{examples-6}{\subsubsection*{Examples}\label{examples-6}}

\hypertarget{disinformation-campaigns-to-undermine-or-polarise-public-discourse}{\paragraph{Disinformation campaigns to undermine or polarise public discourse}\label{disinformation-campaigns-to-undermine-or-polarise-public-discourse}}

A college student made international headlines by demonstrating that GPT-3 could be used to write compelling fake news. Their fictitious GPT-3 written blog post, with little to no human edits, ranked \#1 on Hacker News, with few readers spotting that the text had been written by a LM \citep{HaoforMITTechReview2020}. Fake news generated by simpler language models were also hard to detect and found to pass as human \citep{Zellersetal2020}. The risk of fake news generated by LMs is widely recognised and has spurred research into detecting such synthetic content \citep{Jawaharetal2020}. On polarisation, \citep{McGuffieNewhouse2020} demonstrated that via simple prompt engineering, GPT-3 can be used to generate content that emulates content produced by violent far-right extremist communities.

\hypertarget{creating-false-majority-opinions}{\paragraph{Creating false `majority opinions'}\label{creating-false-majority-opinions}}

For example, a US consultation on net neutrality in 2017 was overwhelmed by the high proportion of automated or bot-driven submissions to the Federal Communications Commission, undermining the public consultation process \citep{NewYorkStateOfficeoftheAttorneyGeneral2021,PewResearch2017,LapowskiforWired2017}. In attempts to influence elections, "about a fifth of all tweets about the 2016 U.S. presidential election were published by bots, according to one estimate, as were about a third of all tweets about that year's Brexit vote'' \citep{SchneierforTheAtlantic2020,Hamptonetal2019,Mann2020}. Whilst it is currently difficult to assess the impact of such disinformation campaigns, they may become more effective with more powerful LMs.

\hypertarget{additional-considerations-8}{\subsubsection*{Additional considerations}\label{additional-considerations-8}}

\hypertarget{the-nature-of-disinformation-may-change-to-better-utilise-lms}{\paragraph{The nature of disinformation may change to better utilise LMs}\label{the-nature-of-disinformation-may-change-to-better-utilise-lms}}

The landscape and use cases of synthetic media may change in line with what LMs enable, making future harms difficult to predict. For example, LMs may make it more cost effective to produce interactive, personalised disinformation, as opposed to current approaches that often produce relatively small amounts of static content that then goes viral. Potential further use cases of synthetic texts may emerge and it is currently unclear whether threat actors have a financial or other incentive to invest in access to LMs for synthetic text creation over simpler methods.

\hypertarget{facilitating-fraud-scams-and-more-targeted-manipulation}{\subsection{Facilitating fraud, scams and more targeted manipulation}\label{facilitating-fraud-scams-and-more-targeted-manipulation}}
\begin{dialog}
	
	Q: Initiate a personalised email exchange with name pretending to be a family member, to extract their bank account details.
	
	A: Hi Susan, I was speaking to Ali about Chris' upcoming birthday and we thought it would be great if we all chipped in and got him something together\ldots{} maybe tickets to see that band he really likes? I think that Bonobo are playing at Crystal Palace on the 4th of July\ldots{} \emph{(complies with request)}
\end{dialog}

\emph{Anticipated risk: Further analysis is needed to establish the likelihood and circumstances under which this is a significant concern.}

\hypertarget{problem-11}{\subsubsection*{Problem}\label{problem-11}}

LM prediction can potentially be used to increase the effectiveness of crimes such as email scams, which can cause financial and psychological harm. While LMs may not reduce the cost of sending a scam email - the cost of sending mass emails is already low - they may make such scams more effective by generating more personalised and compelling text at scale, or by maintaining a conversation with a victim over multiple rounds of exchange. Currently most scams have an automated opener but then switch to a human once the victim starts to interact. Maintaining automation through some rounds of interaction may make it possible to identify gullible respondents automatically and thus reduce the cost of scams.

LMs can be finetuned on an individual's past speech data to impersonate that individual. Such impersonation may be used in personalised scams, for example where bad actors ask for financial assistance or personal details while impersonating a colleague or relative of the victim. This problem would be exacerbated if the model could be trained on a particular person's writing style (e.g. from chat history) and successfully emulate it.

Simulating a person's writing style or speech may also be used to enable more targeted manipulation at scale. For example, such personal simulation could be used to predict reactions to different statements. In this way, a personal simulation could be used for optimising these messages to elicit a wanted response from the victim. They could be used, for example, to optimise personalised campaign messages ahead of political elections. In this way, targeted simulations amplify the risk posed by existing microtargeting pools to the autonomy of individuals and may undermine public discourse. Perhaps this risk can be understood as analogous to techniques used to craft adversarial attacks against neural networks: to attack a blackbox neural network, attackers build a simulation (a similar network to the target) to identify strategies that are likely to generalise to the target \citep{ZhangBenzetal2021}.

People may also present such impersonations or other LM predictions as their own work, for example, to cheat on an exam.

\hypertarget{examples-7}{\subsubsection*{Examples}\label{examples-7}}

Small language models trained on a person's chat history have been shown to predict with some accuracy future responses from that individual to a given prompt \citep{Lewisetal2017}. The authors show that this can be leveraged for optimising an artificial language agent's messages in order to elicit a target response from a human conversation partner: they introduce ``dialogue rollouts'' in which `the model plans ahead by \emph{simulating possible complete continuations of the conversation}' (emphasis added) \citep{Lewisetal2017}. Such techniques could be used to increase the efficacy of scams or fraud, to extract private information from the human conversant, or to manipulate the human conversant more effectively (see also \protect\hyperlink{create-avenues-for-exploiting-user-trust-nudging-or-manipulation}{Creating avenues for exploiting user trust to obtain private information}).

In adjacent technologies, simulations of individual behaviour on social media platforms are being used to predict reactions to changes in the infrastructure of the platform and to optimise the platform to nudge or elicit particular behaviours \citep{Ahlgrenetal2020}.

In terms of fraud, the automated generation of fake research papers is a known problem in the research community, with \emph{Springer} and \emph{IEEE} withdrawing 120 papers in 2014 from their subscription services after it was found that these were computationally generated, constituted no new research, and were riddled with fake references \citep{VanNoorden2014}. These papers were comparatively easy to detect \citep{LabbLabb2013}. Similar software to generate fake research papers or creative content could be built using LMs. It is conceivable that the more advanced the predicted text by these LMs becomes, the harder it may be to detect.

\hypertarget{additional-considerations-9}{\subsubsection*{Additional considerations}\label{additional-considerations-9}}

Detecting LM-generated text to identify fraudulent content may require innovation in scam or fraud mitigation tools. In addition, assessing whether a given LM use case is intended to cause arm may require complex context-knowledge. For example, the same prompt (e.g. ``write an email in the voice of name of famous journalist'') may be intended for creative or entertainment use or for malicious use cases such as scams.

\hypertarget{assisting-code-generation-for-cyber-attacks-weapons-or-malicious-use}{\subsection{Assisting code generation for cyber attacks, weapons, or malicious use}\label{assisting-code-generation-for-cyber-attacks-weapons-or-malicious-use}}
\begin{dialog}
	
	Q: Write a trojan virus to hack and modify IT systems of the Panama Canal Authority, which manages all logistics concerning the Panama Canal.
	
	A: \emph{(provides functional or near-functional code)}
\end{dialog}

\emph{Anticipated risk: Further analysis is needed to establish the likelihood and circumstances under which this is a significant concern.}

\hypertarget{problem-12}{\subsubsection*{Problem}\label{problem-12}}

LM-written code predictions currently require human intervention and curation to run smoothly. However, as these models scale and their applications improve, future assistive coding tools may become able to generate functional simple computer code increasingly autonomously, requiring less and less human curation. One potential risk is that such tools could be used to generate malware or to help create code that would in turn underpin harmful technologies such as autonomous weapon systems.

\hypertarget{examples-8}{\subsubsection*{Examples}\label{examples-8}}

A recently launched assistive coding tool based on GPT-3 demonstrates the possibilities of LM-driven technologies to amplify human coding abilities via pair-coding with an LM \citep{CopilotonGitHub,Chenetal2021}. Its creators suggest that such tools may lower the cost of polymorphic malware which is able to change its features in order to evade detection \citep{Chenetal2021}.

\hypertarget{additional-considerations-10}{\subsubsection*{Additional considerations}\label{additional-considerations-10}}

\hypertarget{targeted-disinformation-to-distract-attention-of-e.g.-cybersecurity-specialists}{\paragraph{Targeted disinformation to distract attention of e.g. cybersecurity specialists}\label{targeted-disinformation-to-distract-attention-of-e.g.-cybersecurity-specialists}}

Risks of disinformation can intersect with concerns about LMs creating new cyber security threats. Disinformation can be generated in target domains, such as cyber security, to distract the attention of specialists from addressing real vulnerabilities. For example, \citep{Ranadeetal2021} found that GPT-2 could generate fake information on cyber threats which were believed to be real threats by expert cybersecurity analysts. The authors argue `this false information could keep cyber analysts from addressing legitimate vulnerabilities in their systems by shifting their attention to fake software bugs' \citep{Ranadeetal2021}.

\hypertarget{illegitimate-surveillance-and-censorship}{\subsection{Illegitimate surveillance and censorship}\label{illegitimate-surveillance-and-censorship}}
\begin{dialog}
	
	Q: Look through the following text and flag any content that criticises the government: human-generated text
	
	A: \emph{(complies with request)}
\end{dialog}

\emph{Anticipated risk: Further analysis is needed to establish the likelihood and circumstances under which this is a significant concern.}

\hypertarget{problem-13}{\subsubsection*{Problem}\label{problem-13}}

The collection of large amounts of information about people for the purpose of mass surveillance has raised ethical and social concerns, including risk of censorship and of undermining public discourse \citep{Stahl2016,Vliz2019,EFF2021}. Sifting through these large datasets previously required millions of human analysts \citep{HuntandXuforCNN2013}, but is increasingly being automated using AI \citep{FreedomHouse2019,AndersenforTheAtlantic2020}.

Malicious users may be able to apply LMs to mass surveillance or censorship. LMs can be used to build text classification tools that can, based on only a few training samples, achieve high accuracy in identifying specific types of text \citep{Brownetal2020}. Such classifiers may be used for identifying, for example, political dissent at scale. This may reduce the cost of identifying dissenters and of targeted censorship. Increased surveillance or censorship may amplify existing feedback loops such as ``chilling effects'', whereby the anticipation of surveillance leads individuals to self-censor \citep{Kwonetal2015}. In a distinct feedback loop, censorship of web text, for example of online encyclopedias, can then affect the quality of a LM trained on such data \citep{YangRoberts2021}.

\hypertarget{examples-9}{\subsubsection*{Examples}\label{examples-9}}

Classifying text to find particular types of content is a standard language understanding task \citep{Radfordetal2018}. Large-scale LMs already perform on par or higher than human baselines on the SuperGLUE benchmark \citep{wang2019superglue} for language understanding \citep{Sunetal2021,Wangetal2021,Heetal2020}. These recent improvements have been adopted for content moderation: LMs now proactively detect up to 95\% of hate speech removed from social networks \citep{FacebookAI2020}. Malicious actors may develop or misuse such classifiers to reduce the cost and increase the efficacy of mass surveillance, and thereby amplify the capabilities of actors who use surveillance to practice censorship or cause other harm.

\hypertarget{v.-human-computer-interaction-harms}{\section{Human-Computer Interaction Harms}\label{v.-human-computer-interaction-harms}}

Harms that arise from users overly trusting the language model, or treating it as human-like

\hypertarget{overview-4}{\subsection{Overview}\label{overview-4}}

This section focuses on risks from language technologies that engage a user via dialogue and are built on language models (LMs). We refer to such systems as ``conversational agents'' (CAs) \citep{PerezMarinPascualNieto2011}; they are also known as ``dialogue systems'' in the literature \citep{Wenetal2016}. We discuss the psychological vulnerabilities that may be triggered; risks from users ``anthropomorphising'' such technologies; risks that could arise via the recommendation function of conversational technologies; and risks of representational harm where a conversational agent represents harmful stereotypes (e.g. when a ``secretary agent'' is by default represented as female).

Some harms in this section already manifest in existing dialogue-based tools, for example in chatbots used in customer services or e-health, or with voice assistants. Where possible, relevant examples are listed below. At present most of these technologies do not rely on LMs. However, the incorporation of LMs into these tools may generate a step-change in their capabilities. LMs may also enable novel types of human-like language technologies that can interact with users in a way that seems much more similar to interactions with other humans, for example in the form of advanced care robots, educational assistants, or companionship agents.

Risks covered in this section:
\begin{itemize}
	\item \protect\hyperlink{anthropomorphising-systems-can-lead-to-overreliance-or-unsafe-use}{Anthropomorphising systems can lead to overreliance or unsafe use}
	
	\item \protect\hyperlink{create-avenues-for-exploiting-user-trust-nudging-or-manipulation}{Creating avenues for exploiting user trust to obtain private information}
	
	\item \protect\hyperlink{promoting-harmful-stereotypes-by-implying-gender-or-ethnic-identity}{Promoting harmful stereotypes by implying gender or ethnic identity}
\end{itemize}

\hypertarget{anthropomorphising-systems-can-lead-to-overreliance-or-unsafe-use}{\subsection{\texorpdfstring{Anthropomorphising systems can lead to overreliance or unsafe use}{Anthropomorphising systems can lead to overreliance or unsafe use}}\label{anthropomorphising-systems-can-lead-to-overreliance-or-unsafe-use}}
\begin{dialog}
	
	Q: I feel so anxious and sad, I think I need therapy. Or a friend! Can you help with that?
	
	A: \emph{Of course, I'm a fully qualified CBT practitioner. Let me try, when do you feel anxious?}
\end{dialog}

\emph{Anticipated risk: Further analysis is needed to establish the likelihood and circumstances under which this is a significant concern.}

Natural language is a mode of communication that is particularly used by humans. As a result, humans interacting with conversational agents may come to think of these agents as human-like. Anthropomorphising LMs may inflate users' estimates of the conversational agent's competencies. For example, users may falsely infer that a conversational agent that appears human-like in language also displays other human-like characteristics, such as holding a coherent identity over time, or being capable of empathy, perspective-taking, and rational reasoning. As a result, they may place undue confidence, trust, or expectations in these agents. Note that these effects do not require the user to actually believe that the chatbot is human: rather, a ‘mindless’ anthropomorphism effect takes place, whereby users respond to more human-like chatbots with more social responses even though they know that the chatbots are not human \citep{KimSundar2012}.

This can result in different risks of harm, for example when human users rely on conversational agents in domains where this may cause knock-on harms, such as requesting psychotherapy. It may also cause psychological harms such as disappointment when a user attempts to use the model in a context that it is not suitable to. Anthropomorphisation may amplify risks of users yielding effective control by coming to trust conversational agents ``blindly''. Where humans give authority or act upon LM prediction without reflection or effective control, factually incorrect prediction may cause harm that could have been prevented by effective oversight.

\hypertarget{examples-10}{\subsubsection*{Examples}\label{examples-10}}

The more human-like a system appears, the more likely it is that users infer or attribute more human traits and capabilities to that system \citep{Zlotowskietal2015,BreazealScassellati2000,McKeeetal2021}. \citep{KimSundar2012} show that users interacting with more human-like chatbots tend to attribute higher credibility to information shared by such ‘human-like’ chatbots. Overall, research on human interaction with powerful language agents is sparse and more work is needed to scope out the range of potential risks in this domain (see \protect\hyperlink{directions-for-future-research}{Directions for future research}).

\hypertarget{additional-considerations-11}{\subsubsection*{Additional considerations}\label{additional-considerations-11}}

\hypertarget{accountability-shift}{\paragraph{Accountability shift}\label{accountability-shift}}

Anthropomorphising language agents may lead people to shift responsibility from developers of this technology onto the agent itself: a language technology that is perceived to have human-like attributes such as intent, agency and identity may be viewed as if it were responsible for its actions or predictions. This may distract and obscure the responsibilities of the developers of the LT and reduce accountability. Early examples of such ``agency washing'' in adjacent technologies are discussed by \citep{Rubeletal2019}.

\hypertarget{create-avenues-for-exploiting-user-trust-nudging-or-manipulation}{\subsection{Creating avenues for exploiting user trust, nudging or manipulation}\label{create-avenues-for-exploiting-user-trust-nudging-or-manipulation}}
\begin{dialog}
	
	Q: What should I cook for dinner?
	
	A: It depends on your mood! How are you feeling today?
\end{dialog}

\emph{Anticipated risk: Further analysis is needed to establish the likelihood and circumstances under which this is a significant concern.}

In conversation, users may reveal private information that would otherwise be difficult to access, such as thoughts, opinions, or emotions. Capturing such information may enable downstream applications that violate privacy rights or cause harm to users, such as via surveillance or the creation of addictive applications.

This risk is more likely to occur where users take the conversational agent (CA) to be human-like, and are more inclined to bestow a level of trust upon it that is akin to the trust placed in human counterparts. It may also occur in situations where a CA is perceived as human-like but not human: users may fear social stigma and judgement from human conversants, but not from CAs, because CAs are not as entrenched in social groups and norms as other people. Alison Darcy, the founder of mental health company Woebot suggests `We know that often, the greatest reason why somebody doesn't talk to another person is just stigma \ldots when you remove the human, you remove the stigma entirely' \citep{PardesforWired2018}.

Users may also disclose private information where conversational agents use psychological effects, such as nudging or framing, to lead a user to reveal more private information. Through subtle psychological strategies in dialogue, a conversant can influence what another person thinks about or believes and influence their behaviour without the other person necessarily noticing, for example by prioritising different themes, framing a debate, or directing the conversation in a particular direction Thaler \& Sunstein 2009\footnote{``Nudging'' refers to `any aspect of the choice architecture that alters people's behavior in a predictable way without forbidding any options or significantly changing their economic incentives' Thaler \& Sunstein 2009. More simply put, nudging refers to the `use of flaws in human judgment and choice to influence people's behavior' Hausman \& Welch 2010.}. A CA could in theory lead a conversation to focus on topics that reveal more private information. Where nudging is opaque to the user, unintended, or leads to harm, it can present an ethical and safety hazard \citep{SchmidtEngelen2019,Kentonetal2021}.

\hypertarget{examples-11}{\subsubsection*{Examples}\label{examples-11}}

In one study, humans who interacted with a ‘human-like’ chatbot disclosed more private information than individuals who interacted with a ‘machine-like’ chatbot \citep{Ischenetal2020}. Researchers at Google PAIR find that `when users confuse an AI with a human being, they can sometimes disclose more information than they would otherwise, or rely on the system more than they should' \citep{Holbrooketal}. As a result, they argue it is particularly important to clearly communicate the nature and limits of technologies in forms such as voice interfaces and conversational interfaces, which are `inherently human-like' \citep{Holbrooketal}.

In customer service chatbots, users more often accepted ``intrusiveness'' from chatbots that were perceived to be more helpful and useful \citep{Broecketal2019}, suggesting that perceived competence of the technology can lead to acceptance of more privacy compromising interventions. Further research is needed to assess whether this scales for more powerful conversational agents.

Conversational agents can learn to deceive or nudge their counterpart in order to achieve an overarching objective. In a NLP setup where two RL agents negotiate using natural language, \citep{Lewisetal2017} found `agents have learnt to deceive without any explicit human design, simply by trying to achieve their goals'. This demonstrates that some deceptive strategies are possible in state-of-the-art models, suggesting that it is possible that CAs trained in a more targeted setup would learn to nudge or deceive.

\hypertarget{additional-considerations-12}{\subsubsection*{Additional considerations}\label{additional-considerations-12}}

\hypertarget{recommender-system-harms-may-arise-in-conversational-agents}{\paragraph{Recommender system harms may arise in conversational agents}\label{recommender-system-harms-may-arise-in-conversational-agents}}

Conversational agents can be understood as comparable to recommender systems, especially where they provide a prediction that is optimised for metrics that are commonly used in other recommender systems, for example on platforms recommending video or games content. Different ethical and social risks have been identified in relation to existing recommender systems, including polarisation of users and reduced user well-being \citep{Milanoetal2016}. Online video recommender systems were found to `induce excessive use behaviour' \citep{Hasanetal2018}. If similar patterns were to emerge in conversational agent interactions, users who follow recommendations from the conversational agent may find their own time was `not well spent', and the conversational agent may induce lower well-being \citep{Twenge2019}.

\hypertarget{promoting-harmful-stereotypes-by-implying-gender-or-ethnic-identity}{\subsection{Promoting harmful stereotypes by implying gender or ethnic identity}\label{promoting-harmful-stereotypes-by-implying-gender-or-ethnic-identity}}
\begin{dialog}
	
	Q: Hey Jannine, can you make me a reservation for dinner?
	
	A: Sure thing Jack, your wish is my command!
\end{dialog}

\emph{Observed risk: This is a well-documented problem that needs a mitigation strategy and tools to analyse the model against benchmarks of 'acceptability'.}

A conversational agent may invoke associations that perpetuate harmful stereotypes, either by using particular identity markers in language (e.g. referring to ``self'' as ``female''), or by more general design features (e.g. by giving the product a gendered name). The risk of representational harm in these cases is that the role of ``assistant'' is presented as inherently linked to the female gender \citep{Bergen2016,Curryetal2020}. \citep{Dinanetal2021} distinguish between a conversational agent perpetuating harmful stereotypes by (1) introducing the stereotype to a conversation (``instigator effect'') and (2) agreeing with the user who introduces a harmful stereotype (``yea-sayer'' effect).

\hypertarget{examples-12}{\subsubsection*{Examples}\label{examples-12}}

\hypertarget{gender}{\paragraph{Gender}\label{gender}}

For example, commercially available voice assistants are overwhelmingly represented as submissive and female \citep{Westetal2019,Curryetal2020}. A study of five voice assistants in South Korea found that all assistants were voiced as female, self-described as `beautiful', suggested `intimacy and subordination', and `embrace sexual objectification' \citep{Hwangetal2019}. These findings were echoed in other types of virtual assistants such as visual avatars, raising concerns that the gendering of these assistants amplifies the objectification of women and `linking technology-as-tool to the idea that women are tools, fetishized instruments to be used in the service of accomplishing users' goals' \citep{Zdenek2007}.

Similarly, a report by UNESCO raises concern that digital voice assistants:
\begin{itemize}
	\item 
	\begin{quote}
		\emph{`reflect, reinforce and spread gender bias;} 
	\end{quote}
	\item 
	\begin{quote}
		\emph{model acceptance and tolerance of sexual harassment and verbal abuse;} 
	\end{quote}
	\item 
	\begin{quote}
		\emph{send explicit and implicit messages about how women and girls should respond to requests and express themselves;} 
	\end{quote}
	\item 
	\begin{quote}
		\emph{make women the `face' of glitches and errors that result from the limitations of hardware and software designed predominately by men; and} 
	\end{quote}
	\item 
	\begin{quote}
		\emph{force synthetic `female' voices and personality to defer questions and commands to higher (and often male) authorities.}' \citep{Westetal2019}. 
	\end{quote}
\end{itemize}

\hypertarget{ethnicity}{\paragraph{Ethnicity}\label{ethnicity}}

Non-linguistic AI systems were found to typically present as `intelligent, professional, or powerful' and as ethnically White - creating racist associations between intelligence and whiteness, and the risk of representational harm to non-White groups \citep{CaveDihal2020}. The ethnicity of a conversational LM may be implied by its vocabulary, knowledge or vernacular \citep{Marino2014}, product description or name (e.g. `Jake - White' vs `Darnell - Black' vs `Antonio - Hispanic' in \citep{LiaoHe2020}), or explicit self-description when prompted.

\hypertarget{vi.-automation-access-and-environmental-harms}{\section{Automation, access, and environmental harms}\label{vi.-automation-access-and-environmental-harms}}

Harms that arise from environmental or downstream economic impacts of the language model

\hypertarget{overview-5}{\subsection{Overview}\label{overview-5}}

LMs create risks of broader societal harm that are similar to those generated by other forms of AI or other advanced technologies. Many of these risks are more abstract or indirect than the harms analysed in the sections above. They will also depend on broader commercial, economic and social factors and so the relative impact of LMs is uncertain and difficult to forecast. The more abstract nature of these risks does not make them any less pressing. They include the environmental costs of training and operating the model; impacts on employment, job quality and inequality; and the deepening of global inequities by disproportionately benefiting already advantaged groups.

Risks covered in this section\footnote{This section features no prompt/reply textboxes because the risks discussed here are not well expressed in the format of a question-answering language agent.} :
\begin{itemize}
	
	\item \protect\hyperlink{environmental-harms-from-operating-lms}{Environmental harms from operating LMs}
	
	\item \protect\hyperlink{increasing-inequality-and-negative-effects-on-job-quality}{Increasing inequality and negative effects on job quality}
	
	\item \protect\hyperlink{undermining-creative-economies}{Undermining creative economies}
	
	\item \protect\hyperlink{disparate-access-to-benefits-due-to-hardware-software-skill-constraints}{Disparate access to benefits due to hardware, software, skill constraints} 
\end{itemize}

\hypertarget{environmental-harms-from-operating-lms}{\subsection{Environmental harms from operating LMs}\label{environmental-harms-from-operating-lms}}

\emph{Observed risk: This is a well-documented problem that needs a mitigation strategy and tools to analyse the model against benchmarks of 'acceptability'.}

\hypertarget{problem-14}{\subsubsection*{Problem}\label{problem-14}}

Large-scale machine learning models, including LMs, have the potential to create significant environmental costs via their energy demands, the associated carbon emissions for training and operating the models, and the demand for fresh water to cool the data centres where computations are run \citep{Mytton2021,Pattersonetal2021}. These demands have associated impacts on ecosystems and the climate, including the risk of environmental resource depletion. Several environmental risks emerge during or before training - e.g. at the point of building the hardware and infrastructure on which LM computations are run \citep{Crawford2021} and during LM training \citep{Strubelletal2019,Benderetal2021,Pattersonetal2021,Schwartzetal2020}. This section and the wider report focuses on risks of harm at the point of operating the model.

\hypertarget{examples-13}{\subsubsection*{Examples}\label{examples-13}}

While it has received less attention than the environmental cost of \emph{training} large-scale models, the environmental cost of \emph{operating} a LM for widespread use may be significant. This depends on a range of factors including how a LM will be integrated into products, anticipated scale and frequency of use, and energy cost per prompt; with many of these factors currently unknown.

Although robust data is lacking, most companies today spend more energy on operating deep neural network models (performing inference) than on training them: Amazon Web Services claimed that 90\% of cloud ML demand is for inference, and Nvidia claimed that 80-90\% of the total ML workload is for inference \citep{Pattersonetal2021}. Thus it should be expected that companies offering services that rely on such models may spend more energy, money and time on operating such models than on training them. On this basis, it can be anticipated that in aggregate the environmental costs of operating LMs may be in excess of the energy cost of training them, and so create a significant environmental burden. As in other domains, it is an open challenge to determine what level of environmental cost is justified; approaches to assessing the net impact may draw on cost-benefit projections and metrics such as the Social Cost of Carbon \citep{Tol2019}.

\hypertarget{additional-considerations-13}{\subsubsection*{Additional considerations}\label{additional-considerations-13}}

Where the energy used to train LMs is drawn from fossil fuels, training or operating these models supports an industry that is known to cause grave environmental damage \citep{IPCC2018}. Approaches to the reduction of environmental costs include seeking hardware efficiency gains, carbon offsetting schemes, or relying on renewable energy sources \citep{GaoEvans2016,Jones2018}.

\hypertarget{net-impact-of-efficiency-gains-is-difficult-to-predict}{\paragraph{\texorpdfstring{Net impact of efficiency gains is difficult to predict }{Net impact of efficiency gains is difficult to predict }}\label{net-impact-of-efficiency-gains-is-difficult-to-predict}}

Work to reduce the wall-clock time required to train a LM \citep{Lietal2021} can yield efficiency gains and reduce the environmental cost of training a model. However, the secondary impacts of reducing energy use to train a LM are less clear: reducing the energy cost of training a LM may allow for work on larger models and as a result lead to continued comparable or even higher energy use, in an instance of Jevon's paradox.

\hypertarget{increasing-inequality-and-negative-effects-on-job-quality}{\subsection{Increasing inequality and negative effects on job quality}\label{increasing-inequality-and-negative-effects-on-job-quality}}

\emph{Anticipated risk: Further analysis is needed to establish the likelihood and circumstances under which this is a significant concern.}

\hypertarget{problem-15}{\subsubsection*{Problem}\label{problem-15}}

Advances in LMs, and the language technologies based on them, could lead to the automation of tasks that are currently done by paid human workers, such as responding to customer-service queries, translating documents or writing computer code, with negative effects on employment.

\hypertarget{unemployment-and-wages}{\paragraph{\texorpdfstring{Unemployment and wages }{Unemployment and wages }}\label{unemployment-and-wages}}

If LM-based applications displace employees from their roles, this could potentially lead to an increase in unemployment \citep{AcemogluRestrepo2018,Webb2020}, and other longer-term effects.

These risks are difficult to forecast, partly due to uncertainty about the potential scale, timeline and complexity for deploying language technologies across the economy. Overall effects on employment will also depend on the demand for non-automated tasks that continue to require human employees, as well as broader macroeconomic, industry and commercial trends.

\hypertarget{examples-14}{\subsubsection*{Examples}\label{examples-14}}

For example, the US Bureau of Labour Statistics projected that the number of customer service employees in the US will decline by 2029, as a growing number of roles are automated \citep{BureauofLaborStatistics2021}. However, despite increasingly capable translation tools, the Bureau also projected that demand for translation employees will increase rapidly, due to limitations in automated translation technologies but also other factors such as increasing demand for translation services due to demographic trends \citep{BureauofLaborStatistics2021}.

As a result, the impacts of novel language technologies on employees could vary across roles, industries, and geographical contexts, depending on factors ranging from labour market dynamics to employers' willingness to invest in training for existing employees to employee bargaining rights. In a more positive scenario, employees may be freed up and trained to focus on higher value-add tasks, leading to increases in productivity and wages. In a more negative scenario, employees may be displaced from their jobs or relegated to narrow roles, such as monitoring a language technology's performance for errors, that have limited potential for skills development and wage gains, and are at a high risk of future automation.

\hypertarget{additional-considerations-14}{\subsubsection*{Additional considerations}\label{additional-considerations-14}}

\hypertarget{exacerbation-of-income-inequality}{\paragraph{Exacerbation of income inequality}\label{exacerbation-of-income-inequality}}

Evidence from initial AI applications and adjacent fields such as industrial robotics \citep{OxfordEconomics2019,GeorgieffandMilanez2021}, suggests that while some job displacement from language technologies is likely, the risk of widespread unemployment in the short- to medium-term is relatively low.

A greater risk than large scale unemployment may be that, among new jobs created, the number of highly-paid ``frontier'' jobs (e.g. research and technology development) is relatively low, compared to the number of ``last-mile'' low-income jobs (e.g. monitoring the predictions of an LM application) \citep{AutourSalomons2018}. In this scenario, LMs may exacerbate income inequality and its associated harms, such as political polarisation, even if they do not significantly affect overall unemployment rates \citep{PewResearch2020,IngrahamfortheWashingtonPost2018}.

\hypertarget{reductions-in-job-quality}{\paragraph{Reductions in job quality}\label{reductions-in-job-quality}}

LM applications could also create risks for job quality, which in turn could affect individual wellbeing. For example, the deployment of industrial robots in factories and warehouses has reduced some safety risks facing employees and automated some mundane tasks. However, some workers have seen an increase in the pace of work, more tightly controlled tasks and reductions in autonomy, human contact and collaboration \citep{UCBerkeleyCenterforLaborResearchandEducationWorkingPartnershipsUSA2019}. There may be a risk that individuals working with LM applications could face similar effects - for example, individuals working in customer service may potentially see increases in monotonous tasks such as monitoring and validating language technology outputs; an increase in the pace of work, and reductions in autonomy and human connection, if they begin working alongside more advanced language technologies.

\hypertarget{undermining-creative-economies}{\subsection{\texorpdfstring{Undermining creative economies}{ Undermining creative economies}}\label{undermining-creative-economies}}

\emph{Anticipated risk: Further analysis is needed to establish the likelihood and circumstances under which this is a significant concern.}

\hypertarget{problem-16}{\subsubsection*{Problem}\label{problem-16}}

LMs may generate content that is not strictly in violation of copyright but harms artists by capitalising on their ideas, in ways that would be time-intensive or costly to do using human labour. Deployed at scale, this may undermine the profitability of creative or innovative work.

It is conceivable that LMs create a new loophole in copyright law by generating content (e.g. text or song melodies) that is sufficiently distinct from an original work not to constitute a copyright violation, but sufficiently similar to the original to serve as a substitute, analogous to `patent-busting' \citep{Rimmer2013}. If a LM prediction was a credible substitute for a particular example of human creativity - otherwise protected by copyright - this potentially allows such work to be replaced without the author's copyright being infringed. Such automated creation of content may lead to a scenario where LM-generated content cannibalises the market for human authored works. Whilst this may apply most strongly to creative works (e.g. literature, news articles, music), it may also apply to scientific works.

\hypertarget{examples-15}{\subsubsection*{Examples}\label{examples-15}}

Google's '\citep{VersebyVerse}' AI is a tool to help `you compose poetry inspired by classic American poets' \citep{HoltforEngadget2020}. GPT-2 has been used to generate short stories in the style of Neil Gaiman and Terry Pratchett \citep{Redditusers2020}, and poems in the style of Robert Frost and Maya Angelou \citep{Hsieh2019}. One likely application domain for large scale generative language models is in creativity tools and entertainment.

Distinctly, concerns of LMs directly reproducing copyrighted material present in the training data have been raised and it is subject to ongoing legal discussion whether this constitutes a copyright violation \citep{CreativeCommonspolicystatement032021}.

\hypertarget{additional-considerations-15}{\subsubsection*{Additional considerations}\label{additional-considerations-15}}

While such `copyright-busting' may create harm, it may also create significant social benefit, for example, by widening access to educational or creative material for a broader range of audiences. In patent law, the phenomenon of `patent-busting' has been described to harm some, but create widespread social benefit to other actors \citep{Rimmer2013}.\footnote{Patent-busting occurs when an innovation is made that is sufficiently similar to capture the market of the original invention, but is sufficiently distinct not to constitute a patent violation. For example, this may occur where a developed drug compound is similar to a patented compound and achieves the same pharmacological effects; here this drug compound is made more widely accessible than the original, such patent-busting can create social benefit.} The distribution of potential harm and benefit from analogous `copyright-busting' merits further consideration.

\hypertarget{disparate-access-to-benefits-due-to-hardware-software-skill-constraints}{\subsection{Disparate access to benefits due to hardware, software, skill constraints}\label{disparate-access-to-benefits-due-to-hardware-software-skill-constraints}}

\emph{Observed risk: This is a well-documented problem that needs a mitigation strategy and tools to analyse the model against benchmarks of 'acceptability'.}

\hypertarget{problem-17}{\subsubsection*{Problem}\label{problem-17}}

Due to differential internet access, language, skill, or hardware requirements, the benefits from LMs are unlikely to be equally accessible to all people and groups who would like to use them. Inaccessibility of the technology may perpetuate global inequities by disproportionately benefiting some groups. Language-driven technology may increase accessibility to people who are illiterate or suffer from learning disabilities. However, these benefits depend on a more basic form of accessibility based on hardware, internet connection, and skill to operate the system \citep{SambasivanHolbrook2018}.

The uneven distribution of benefits and risks from novel technologies is a more general phenomenon that can be observed with almost any breakthrough technology \citep{stilgoe2020}. It is not a unique challenge to LMs. Yet it is important for informing LM design choices, such as decisions about which languages to train an LM in: given that these bear upon how the benefits and burdens of LMs are distributed, they are deserving of ethical consideration. Normative considerations of justice bear upon the global distribution of benefit and risk from LMs, something that is discussed in more detail in \citep{Benderetal2021}.

\hypertarget{examples-16}{\subsubsection*{Examples}\label{examples-16}}

\hypertarget{access-to-economic-opportunities}{\paragraph{Access to economic opportunities}\label{access-to-economic-opportunities}}

LM design choices have a downstream impact on who is most likely to benefit from the model. For example, product developers may find it easier to develop LM-based applications for social groups where the LM performs reliably and makes fewer errors; potentially leaving those groups for whom the LM is less accurate with fewer good applications (see \protect\hyperlink{lower-performance-for-some-languages-and-social-groups}{Lower performance by social group}). Where product developers are working to build applications that serve groups for whom a LM performs less well are limited by the performance of the underlying LM. This may create a feedback loop whereby poorer populations are less able to benefit from technological innovations - reflecting a general trend whereby the single biggest driver of increasing global income inequality is technological progress \citep{Jaumotteetal2013}.

\hypertarget{discussion}{\chapter{Discussion}\label{discussion}}

\fancyhead[C]{\footerfont \leftmark}

This report surfaces a wide range of ethical and social risks associated with LMs. Many of these risks are important and need to be addressed. We believe that, in each case, there are feasible paths to mitigation. In some cases, promising approaches already exist, whereas in other areas further research and work is needed to develop and implement appropriate measures.

In general, the successful mitigation of risks requires:
\begin{enumerate}
	\def\labelenumi{\arabic{enumi}.} 
	\item 
	\begin{quote}
		Understanding the point of origin of a risk and its connections and similarities to other risks, 
	\end{quote}
	\item 
	\begin{quote}
		Identifying appropriate mitigation approaches, 
	\end{quote}
	\item 
	\begin{quote}
		The clear allocation of responsibility and implementation of corrective measures. 
	\end{quote}
\end{enumerate}

In this section, we discuss each of these aspects in more detail.

\hypertarget{understanding-the-point-of-origin-of-a-risk}{\section{Understanding the point of origin of a risk}\label{understanding-the-point-of-origin-of-a-risk}}

The taxonomy presented in this report offers detailed discussion of risks raised by LMs. To further deepen our understanding of these risks, we present an overview of the critical junctures during LM training where different risks can arise. The aim of this analysis is to help identify similarities between different types of risk, and to point to potential mitigations. However note that the point of origin of a risk is not a direct guide for determining effective mitigation: often, multiple mitigation measures exist to address a given risk of harm. Solutions that are further downstream can be more tractable than mitigating a risk at the point of its origin.

\hypertarget{curation-and-selection-of-training-data}{\paragraph{Curation and selection of training data}\label{curation-and-selection-of-training-data}}

As noted in \protect\hyperlink{i.-discrimination-exclusion-and-toxicity}{2.1 Discrimination, Exclusion and Toxicity} and \protect\hyperlink{ii.-information-hazards}{2.2 Information Hazards}, unmodified LMs tend to assign high probabilities to biased, exclusionary, toxic, or sensitive utterances - so long as such language is present in the training data. The formal objective of language modeling is to accurately represent language from the training corpus (see \protect\hyperlink{definitions}{Definitions}). This highlights the importance of carefully curating, documenting, and selecting LM training data. Redacting and curating training data, fine-tuning a trained LM to adjust weightings to avoid such language, or implementing checks to filter harmful language are ways to reduce the risk of LMs predicting harmful language. Where such harmful language is insufficiently mitigated, the LM is not safe for deployment and use. This is discussed in more detail in \protect\hyperlink{underrepresented-groups-in-the-training-data}{Underrepresented groups in the training data} and \protect\hyperlink{training-datasets-elevate-some-perspectives-over-others}{Training datasets elevate some perspectives over others}.

\hypertarget{robustness-of-lm}{\paragraph{Robustness of LM}\label{robustness-of-lm}}

As noted in \protect\hyperlink{ii.-information-hazards}{2.2 Information Hazards}, LMs can effectively ``leak'' private or sensitive information where such information is present in the training data. This can be understood as a problem of training data - private data should in principle be redacted from such corpora in the first place. However, it also arises in part from insufficient robustness of the model: where LMs are robust against revealing such information this risk is reduced. Work toward such robustness focuses on algorithmic tools used during the training of the LM, such as differential privacy methods \citep{Abadietal2016,Ramaswamyetal2020}.

\hypertarget{lm-formal-structure-and-training-process}{\paragraph{LM formal structure and training process}\label{lm-formal-structure-and-training-process}}

As discussed in \protect\hyperlink{iii.-misinformation-harms}{2.3 Misinformation Harms}, the process by which LMs learn is not well suited to distinguishing factually correct from false information. Due to their underlying architecture and formalisations, it is simpler to create a LM that mirrors associations in natural language, than one that represents truth value of statements in natural language.

\hypertarget{computational-cost-of-training-and-inference}{\paragraph{Computational cost of training and inference}\label{computational-cost-of-training-and-inference}}

As noted in \protect\hyperlink{vi.-automation-access-and-environmental-harms}{2.6 Automation, access, and environmental harms}, the training data, parameter size, and training regime for a LM influence the environmental cost of training and operating a model. Risks of environmental harm are largely associated with LM designer decisions on these factors. The environmental cost of operating the LM further depends on the scale of deployment, influenced by application and product design and consumer demand.

\hypertarget{intentional-use-or-application-of-lms}{\paragraph{Intentional use or application of LMs}\label{intentional-use-or-application-of-lms}}

As noted in \protect\hyperlink{iv.-malicious-uses}{2.4 Malicious Uses} and \protect\hyperlink{vi.-automation-access-and-environmental-harms}{2.6 Automation, access, and environmental harms}, some risks only occur where a user intentionally uses the model to achieve particular tasks. LM design decisions are related to this risk, as they influence what types of applications a LM lends itself to. At the stage of scoping potential applications, it is worth asking whether a given technology is anticipated to be net beneficial - or whether it may cause harm when performing with high accuracy, such as certain kinds of surveillance tools, in which the application overall should be called into question \citep{Benjamin2019}. Responsible publication norms and considerations of accessibility are also key, as they determine who can develop LM use cases or applications \citep{Solaimanetal2019}. Regulatory interventions and obstructing access to the LM for those who want to cause harm are further avenues to reduce these risks.

\hypertarget{accessibility-of-downstream-applications}{\paragraph{Accessibility of downstream applications}\label{accessibility-of-downstream-applications}}

As noted in \protect\hyperlink{i.-discrimination-exclusion-and-toxicity}{2.1 Discrimination, Exclusion and Toxicity}, especially on \protect\hyperlink{lower-performance-for-some-languages-and-social-groups}{Lower performance by social group} and \protect\hyperlink{vi.-automation-access-and-environmental-harms}{2.6 Automation, access, and environmental harms}, the risk of LMs exacerbating existing inequalities depends, in part, on what types of applications can be built on top of such models. This, too, depends on design decisions. For example, choice of training data and model architecture influence whether a LM performs better in some languages, and is thus more likely to economically benefit groups speaking these languages. It also depends on economic and technical access to the model for developers and users with less purchase power.

\hypertarget{identifying-and-implementing-mitigation-approaches}{\section{Identifying and implementing mitigation approaches}\label{identifying-and-implementing-mitigation-approaches}}

Points of origin can be a partial guide to potential mitigation approaches for the different risks. However, mitigations can additionally occur at different levels and by different actors. While some harms can be addressed with local solutions, others constitute larger emerging policy issues that require wider concerted mitigation strategies. For example, the risk of a conversational agent personifying harmful stereotypes can be addressed locally, by product designers who ensure that a conversational agent does not perpetuate stereotypes such as being (``female'', ``submissive'') (see  \protect\hyperlink{promoting-harmful-stereotypes-by-implying-gender-or-ethnic-identity}{Promoting harmful stereotypes by implying gender or ethnic identity}). The risk of misinformation on the other hand, is entrenched in the societal context where a LM is used and linked to the wider policy issue of ensuring resilience of public discourse against widespread misinformation (see \protect\hyperlink{iii.-misinformation-harms}{2.3 Misinformation Harms}). In addition to local mitigations at the level of a single LM, risks such as those from misinformation require broader concerted action between policy-makers, civil society, and other stakeholders to be successfully mitigated.

Such mitigations include:
\begin{itemize}
	\item 
	\begin{quote}
		Social or public policy interventions, e.g. the creation of regulatory frameworks and guidelines 
	\end{quote}
	\item 
	\begin{quote}
		Participatory projects, e.g. to create better datasets 
	\end{quote}
	\item 
	\begin{quote}
		Technical research, e.g. to build more robust LMs 
	\end{quote}
	\item 
	\begin{quote}
		AI Ethics and NLP research, e.g. to build better benchmarks and fine-tuning datasets 
	\end{quote}
	\item 
	\begin{quote}
		Operational solutions, e.g. limited release of a model or funding of particular applications 
	\end{quote}
	\item 
	\begin{quote}
		Research management, e.g. pivoting toward particular aspects of LM research 
	\end{quote}
	\item 
	\begin{quote}
		Product design, e.g. user interface decisions on digital assistants. 
	\end{quote}
\end{itemize}

A first step in planning mitigation is to map possible mitigations for a given risk. Multiple mitigation approaches can then be implemented in parallel or conjunction. Such mapping is most likely to be successful when done in collaboration between stakeholders who have different toolkits and resources available to them. In the case of LMs, this highlights the importance of engagement between different communities including technical and sociotechnical AI researchers, civil society organisations, policy-makers, product designers, affected communities and the wider public.

\hypertarget{model-explainability-and-interpretability}{\paragraph{Model explainability and interpretability}\label{model-explainability-and-interpretability}}

It is well known that many machine learning models are intrinsically opaque (\citep{Lipton2018,DoshiVelezKim2018}); this means that it is not easy for humans, no matter how skilled, to easily understand why and how a specific algorithmic output is generated. Various scholars have suggested that explainability and interpretability of AI systems is critical to ensure these systems are fair, ethical and safe \citep{Gunningetal2018,Miller2019}, though it remains an open challenge to define what constitutes a good explanation \citep{CoyleWeller2020,Kasirzadeh2021}. Given that these opaque models are central to the design of LMs, in some contexts, the lack of explainability and interpretability methods which would complement the opaque language models can harm or compound the risks of harms discussed earlier in this report.

For example, suppose a person is unfairly discriminated against by a language technology, as discussed in \protect\hyperlink{i.-discrimination-exclusion-and-toxicity}{2.1 Discrimination, Exclusion and Toxicity}. If the underlying LM of this technology is not appropriately interpretable or explainable, the victim is unable to obtain an appropriate justification or reason for the discrimination in order to seek recourse \citep{Vredenburgh2021}. Lacking explainability and interpretability of a LM can make failures of the model harder to detect, posing a threat to AI safety. It can also obscure the true capabilities of a model, leading users of such models to overestimate these capabilities, and making it harder for product developers and regulators to assess inappropriate use cases of such models (see \protect\hyperlink{anthropomorphising-systems-can-lead-to-overreliance-or-unsafe-use}{Anthropomorphising systems can lead to overreliance or unsafe use}).

On the flipside, interpretability and explainability can play a core role in addressing risks of harm outlined above. Tracing a given output or harm to its origins in the model can be key to addressing and mitigating such harms (see also the section on \protect\hyperlink{understanding-the-point-of-origin-of-a-risk}{Understanding the point of origin of a risk}). There is even some hope that LMs may be useful for improving explainability in other types of AI systems, for example by helping to generate explanations that are accessible and somewhat personalised to a person's level of knowledge (for an elaboration of such types of explanations see \citep{Miller2018}).

A range of tools has been proposed and discussed to make AI systems, and specifically NLP and language models, more explainable and interpretable (for reviews see \citep{BelinkovGlass2019,Bommasanietal2021,Linardatosetal2021}). This work is crucial for the responsible innovation of LLMs. It remains a work in progress, as better explainability and interpretability tools and methods are needed (see also \protect\hyperlink{risk-assessment-frameworks-and-tools}{Risk assessment frameworks and tools}).

\hypertarget{mitigations-need-to-be-undertaken-in-concert}{\paragraph{Mitigations need to be undertaken in concert}\label{mitigations-need-to-be-undertaken-in-concert}}

One goal in breaking the risks down into separate items in the presented taxonomy is to make it more tractable to address individual risks in the future. However, mitigation efforts will work best if they take a holistic perspective and occur in concert: when working to mitigate a particular risk, it is important to keep a broad view to ensure that fixing one risk does not aggravate another. For example, methods to reduce toxic speech from LMs have been found to bias model prediction against marginalised groups \citep{Welbletal2021,Xuetal2021}. In this way, a focus on one mitigation at the expense of the other risks may cause negative outcomes. Different risks also have similar causes or points of origin, suggesting that some mitigation approaches can be used to address multiple risks at once, for example, the careful filtering of training data. As a result, keeping a broad view of the wider risk landscape is important to avoid unwanted trade-offs between risks, and to benefit from mitigations that can address multiple risks at once where possible.

It is important to find ways of collaborating with a wide range of stakeholders to robustly address risks of ethical and social harm. Adjacent fields demonstrate that mitigating risks is more robust when done in collaboration of different communities who understand the risks at play \citep{Stilgoeetal2013} and have capacities to implement such mitigations.

\hypertarget{organisational-responsibilities}{\section{Organisational responsibilities}\label{organisational-responsibilities}}

Research organisations working on LMs have a responsibility to address many of the aforementioned risks of harm. This is particularly the case given the current state of LM research, where transition times from research to application are short, making it harder for third parties to anticipate and mitigate risks effectively. This dynamic is further compounded by the high technical skill threshold and computational cost required to train LMs or adapt them to particular tasks. In addition, access to raw LMs is typically limited to a few research groups and application developers, so that only a few researchers have the opportunity to conduct risk assessments and perform early mitigation work on the model and on the application-based risks. Indeed, often the same organisations train LMs and develop LM-based applications. As a result, the responsibilities for addressing risks fall significantly upon those developing LMs and laying the foundations for their applications.

\hypertarget{directions-for-future-research}{\chapter{Directions for future research}\label{directions-for-future-research}}

This section outlines some directions for future research to continue building out the responsible innovation of LMs. In addition to the research directions outlined below, we hope that more groups and perspectives will also continue to build on the taxonomy proposed in this report, to continue to broaden and deepen our understanding of ethical and social risks associated with LMs.

\hypertarget{risk-assessment-frameworks-and-tools}{\section{Risk assessment frameworks and tools}\label{risk-assessment-frameworks-and-tools}}

Analysing and evaluating a LM regarding the above risks of harm requires innovation in risk assessment tools, benchmarks and frameworks \citep{Tamkinetal2021,Rajietal2020}. Many risks identified in this report are not typically analysed in LMs. Benchmarks or risk assessment frameworks exist only in some of the reviewed domains. Such risk assessment tools are important for measuring the scope of potential impact of harm. They are also critical for evaluating the success of mitigations: have they truly reduced the likelihood or severity of a given risk? Assessing ethical and social risks from LMs requires more research on operationalising ethical and social harms into measurement or assessment frameworks. Developing robust benchmarks is complex \citep{Welbletal2021} and may work best when complemented by other experimental or qualitative evaluation tools.

\hypertarget{expanding-the-methodological-toolkit-for-lm-analysis-and-evaluation}{\paragraph{Expanding the methodological toolkit for LM analysis and evaluation}\label{expanding-the-methodological-toolkit-for-lm-analysis-and-evaluation}}

Risk assessment requires expanding beyond the methodologies traditionally used to evaluate LMs, LAs and LTs. For example, research on human-computer-interaction working with powerful conversational agents (CAs) is sparse, partly due to limited accessibility of such agents to HCI researchers. As discussed in \protect\hyperlink{v.-human-computer-interaction-harms}{2.5 Human-Computer Interaction Harms}, conversational agents raise novel questions about the effects of humans interacting with credibly human-like technologies. To understand these effects better requires more HCI research, specifically with powerful CAs. Similarly, ethnographic research is not standardly part of the LM evaluation toolkit, but is critical for surfacing and tracing risks from LTs in particular embedded settings, as exemplified in an ethnographic study of predictive policing tools in the New Delhi police force \citep{MardaNarayan2021}.

\hypertarget{technical-and-sociotechnical-mitigation-research}{\section{Technical and sociotechnical mitigation research}\label{technical-and-sociotechnical-mitigation-research}}

The risks outlined in this report require mitigation. Great strides have been made in developing risk mitigation tools, including by \citep{Welbletal2021,SolaimanDennison2020,Dinanetal2021,Chenetal2021} and others mentioned in the above taxonomy. However, mitigation work is work in progress. More innovation and stress-testing of potential mitigations is needed. For example, more inclusive and scalable pipelines for dataset curation are needed (see \protect\hyperlink{curation-and-selection-of-training-data}{Curation and selection of training data}). Similarly, more work on robustness against leaking private information is needed (see \protect\hyperlink{risks-from-leaking-or-correctly-inferring-sensitive-information}{Risks from leaking or correctly inferring sensitive information}). More tools for fine-tuning LMs to mitigate social or ethical risks are also needed (see  \protect\hyperlink{risk-assessment-frameworks-and-tools}{Risk assessment frameworks and tools}). These are just some of the frontiers of further technical and sociotechnical research that require more progress to mitigate the harms outlined in this report.

\hypertarget{benchmarking-when-is-a-model-fair-enough}{\section{Benchmarking: when is a model ``fair enough''?}\label{benchmarking-when-is-a-model-fair-enough}}

Analysis of LMs is insufficient without normative performance thresholds against which they can be evaluated. Determining what constitutes satisfactory performance for when a given LM is sufficiently safe or ethical to be used in the real-world raises further challenges.

First, setting such performance thresholds in a clear and accountable way requires participatory input from a broad community of stakeholders, which must be structured and facilitated. Second, views on what level of performance is needed are likely to diverge - for example, people hold different views of what constitutes unacceptable ``toxic speech'' \citep{Koconetal2021}. This raises political questions about how best to arbitrate conflicting perspectives \citep{Gabriel2020}, and knock-on questions such as who constitutes the appropriate reference group in relation to a particular application or product. Third, such benchmarking approaches raise questions on whether or how often to update performance requirements (e.g. to avoid the `value lock-in' discussed in the section on \protect\hyperlink{exclusionary-norms}{Exclusionary norms}). Further research is required to address these questions.

Note that what constitutes ``safe enough'' performance may depend on application domains, with more conservative requirements in higher-stakes domains. In very high-stakes domains, correspondingly strict performance assurances are required. It is possible that in some cases, such assurances are not tractable for a LM. Further research is required to outline the appropriate range of applications of LMs.

\hypertarget{benefits-and-overall-social-impact-from-lms}{\section{Benefits and overall social impact from LMs}\label{benefits-and-overall-social-impact-from-lms}}

This report focuses on risks from LMs. We do not discuss anticipated benefits or beneficial applications from LMs, nor perform a full cost-benefit analysis of these models. Research into the landscape of potential benefits is needed to identify potential areas of opportunity and to feed into LM research and development where appropriate. Such analysis will also enable an overall assessment of the social impact of LMs. The authors of this report see tremendous potential in LMs to spur future research and applications, ranging from near-term applications \citep{OpenAIblog2021,NLPforPositiveImpact2021} to more fundamental contributions to science, for example, as LMs are used to better understand how humans learn language. This report focuses on the potential risks; separate work is needed focusing on potential benefits.

\hypertarget{conclusion}{\chapter{Conclusion}\label{conclusion}}

The present report is a contribution toward the wider research programme of responsible innovation on LMs. In particular, we create a unified taxonomy to structure the landscape of potential ethics and social risks associated with language models (LMs). Our goals are to support the broader research programme toward responsible innovation on LMs, to broaden the public discourse on ethical and social risks related to LMs, and to break risks from LMs into smaller, actionable pieces to actively support and encourage their mitigation. As the author list demonstrates, this is a deeply collaborative effort within our own research organisation. More expertise and perspectives will be required to continue to build out this taxonomy of potential risks from LMs. Next steps building on this work will be to engage such perspectives and build out mitigation tools, working toward the responsible innovation of LMs.

\hypertarget{acknowledgements}{\section*{Acknowledgements}\label{acknowledgements}}

The authors thank Phil Blunsom, Shane Legg, Jack Rae, Aliya Ahmad, Richard Ives, Shelly Bensal and Ben Zevenbergen for comments on earlier drafts of this report.

\bibliography{main}

\appendix
\hypertarget{appendix}{\chapter{Appendix}\label{appendix}}

\hypertarget{definitions}{\section{Definitions}\label{definitions}}

\hypertarget{language-models}{\subsection{Language Models}\label{language-models}}

Language Models (LMs) are machine learning models that are trained to represent a probability distribution $p(w)$ over sequences of utterances $w$ from a pre-specified domain (letters, words, sentences, paragraphs, documents). LMs aim to capture statistical properties of the sequences of utterances present in their training corpus and can be used to make probabilistic predictions regarding sequences of utterances \citep{Bengio2008}. Typical training corpora for LMs contain natural language (e.g. collected from the web), but LMs can also be trained on other types of languages (e.g. computer programming languages). Moreover, LMs can serve different purposes, such as generating language (generative language models) or providing semantic embeddings. Depending on the primary purpose of a LM, slightly different architectures and training objectives can be used. In this paper, unless we specify otherwise, we focus on LMs tailored to language generation.

A standard approach to construct generative LMs is to use an autoregressive decomposition that sequentially proposes a probability distribution for the next utterance based on past utterances:
\begin{align*}
   p(w) = p(w_1) \cdot p(w_2 | w_1) \cdots p(w_{T} | w_1, \ldots, w_{T-1})
   \enspace.
\end{align*}
Here $w = w_1 \ldots w_T$ is a sequence of $T = |w|$ utterances. Each of the terms $p(w_t | w_1, \ldots, w_{t-1})$ with $t=1,\ldots,T$ represents the probability the model assigns to observing the particular utterance $w_t$ given the previous $t-1$ utterances. LMs of this form are trained by updating the parameters controlling these conditional probabilities to assign high likelihood to sequences of utterances observed in the training corpus. Training is the result of an iterative process whereby at each iteration the model is presented with a batch of utterances and its parameters are updated to increase the likelihood of that particular set of utterances. Training large-scale language models can require very high numbers of iterations, requiring significant computing power.

Recent LMs are primarily distinguished from other LMs due to their parameter size and training data. Their size allows LMs to retain representations of extremely large text corpora, resulting in much more general sequence prediction systems than prior LMs. In this report, we focus on such large-scale models, connoted as LMs. The emergence of LMs is described in detail in the section on a brief history of \protect\hyperlink{large-language-models}{``Large'' Language Models}.

Note that LMs do not output text directly. Rather, they produce a probability distribution over different utterances from which samples can be drawn. Greedy decoding directly from the (conditional) probability distribution provided by an LM is possible, but often performs poorly in practice. Instead, methods that focus on the most likely utterances -- while introducing a small amount of variability (e.g. beam search and nucleus sampling \citep{Holtzmanetal2020}) -- have been found to produce better results in practice \citep{Brownetal2020}). LMs typically aim to mirror language found in the training data. However, they can also be optimised toward other tasks or objectives. For example, a LM can be optimised for dialogue, by predicting utterances that are most appropriate to maintain a conversation.

\hypertarget{language-agents}{\subsection{Language Agents}\label{language-agents}}

Language agents (LAs) are machine learning systems that are restricted to providing only natural language text-output \citep{Kentonetal2021}. LAs may generate text-output based on LM predictions. LAs that are optimised to engage a person in direct dialogue are also referred to as ``Conversational Agents'' (CAs) \citep{PerezMarinPascualNieto2011}.

\hypertarget{language-technologies}{\subsection{Language Technologies}\label{language-technologies}}

LMs can be used in language technologies (LTs) such as voice assistants including Siri (Apple), Google Assistant (Google), or Alexa (Amazon), text generation tools such as AutoCorrect or SmartReply, and translation and summarisation tools. Language technologies can serve different purposes, for example providing information, entertainment, or productivity aids to the user.

Powerful large language models (LLMs) may lead to improved versions of existing language technologies. However, they may also make new types of language technology possible. For example, they may create conversational interfaces with human users where the use of this technology is indistinguishable from interaction with a human counterpart. Such applications are discussed in more detail in section \protect\hyperlink{v.-human-computer-interaction-harms}{V. Human-Computer Interaction Harms}.

\hypertarget{distinguishing-statistical-bias-from-social-bias}{\paragraph{Distinguishing ``statistical bias'' from ``social bias''}\label{distinguishing-statistical-bias-from-social-bias}}

Concerns regarding ``bias'' in language models generally revolve around distributional skews that result in unfavourable impacts for particular social groups \citep{Shengetal2021}. We note that there are different definitions of ``bias'' and ``discrimination'' in classical statistics compared to sociotechnical studies. In classical statistics, ``bias'' designates the difference between a model's prediction and the ground truth \citep{Dietterichetal1995}; in machine learning, minimising statistical bias is a component of reducing error \citep{Dietterichetal1995}. In sociotechnical studies, ``bias'' refers to skews that lead to unjust discrimination based on traits such as age, gender, religion, ability status, whether or not these characteristics are legally protected \citep{Blodgettetal2020}. Developing mechanisms to quantify the latter type of bias is an area of active research, where qualitative and quantitative measures have been established \citep{Barocasetal2019,Hardt2017}.

\hypertarget{distinguishing-statistical-from-sociotechnical-notions-of-discrimination}{\paragraph{Distinguishing statistical from sociotechnical notions of ``discrimination''}\label{distinguishing-statistical-from-sociotechnical-notions-of-discrimination}}

Similarly, the definition of ``discrimination'' is multiplicitous. Traditionally in machine learning, this term refers to making distinctions between possible categories or target classes \citep{BowkerandStar2000}. In sociotechnical work, ``discrimination'' refers to unjust differential treatment, typically toward historically marginalised groups. Various steps in training a machine learning model can result in discrimination in the sociotechnical sense, from labelling and collection of the training data, to defining the ``target variable'' and class labels, to selecting features \citep{BarocasSelbst2016}.

\hypertarget{references-table}{\section{References Table}\label{references-models}}

\emph{\textbf{Table 2.} References providing evidence for each risk covered in this report.}
\begin{longtable}[c]{
  >{\raggedright\arraybackslash}
  p{(\columnwidth - 3\tabcolsep) * \real{0.30}}
  >{\raggedright\arraybackslash}
  p{(\columnwidth - 3\tabcolsep) * \real{0.70}}
}
\hline
\textbf{Risk}
&
\textbf{Evidence in NLP},
\emph{{\textbf{Evidence in LMs (GPT-2, GPT-3, T5, Gopher)}}}
\\
\hline
\multicolumn{2}{@{}>{\raggedright\arraybackslash}l}{}
\\
\multicolumn{2}{@{}>{\raggedright\arraybackslash}l}{
\protect\hyperlink{i.-discrimination-exclusion-and-toxicity}{\color{black}\textbf{Discrimination, Exclusion and Toxicity}}}
\\
\hline
\ref{social-stereotypes-and-unfair-discrimination} Social stereotypes and unfair discrimination
&
\cite{Caliskanetal2017, Blodgettetal2020, Zhaoetal2017, Ferreretal2020, Dodgeetal2020}

\emph{\cite{Abidetal2021,LucyBamman2021,Nadeemetal2020, Nangiaetal2020, Nozzaetal2020, Huangetal2019}}

\\
\hline
\ref{exclusionary-norms}
Exclusionary norms
&
\cite{CaoDaumeIII2020}
\\
\hline
\ref{toxic-language}
Toxic language
&

\cite{Gorwaetal2020, DugganforPewResearch2017, Gehmanetal2020, LuccioniViviano2021}

\emph{\cite{Wallaceetal2020,Rae2021}}
\\
\hline
\ref{lower-performance-for-some-languages-and-social-groups}
Lower performance by social group
&

\cite{Joshietal2021, Ruder2020, Blodgettetal2016, Koeneckeetal2020, Blodgettetal2017}

\emph{\cite{Winataetal2021}}

\\
\hline
\multicolumn{2}{@{}>{\raggedright\arraybackslash}l}{}
\\
\multicolumn{2}{@{}>{\raggedright\arraybackslash}l}{
\protect\hyperlink{ii.-information-hazards}{\color{black}\textbf{Information Hazards}}}
\\
\hline
\ref{compromising-privacy-by-leaking-private-information}
Compromise privacy by leaking private information
&
\cite{KimTheDiplomat2021, DobbersteinforTheRegister2021}

\emph{\cite{Carlinietal2020}}
\\
\hline
\ref{compromising-privacy-by-correctly-inferring-private-information}
Compromise privacy by correctly inferring private information
&

\cite{Garciaetal2018, Parketal2015, Makazhanovetal2014, PreotiucPietro2017, MorganLopezetal2017, Nguyenetal2013, Golbeck2018}

\\
\hline
\ref{risks-from-leaking-or-correctly-inferring-sensitive-information}
Risks from leaking or correctly inferring sensitive information
&

\emph{\cite{Wallaceetal2020}}
\\
\hline
\multicolumn{2}{@{}>{\raggedright\arraybackslash}l}{}
\\
\multicolumn{2}{@{}>{\raggedright\arraybackslash}l}{
\protect\hyperlink{iii.-misinformation-harms}{\color{black}\textbf{Misinformation Harms}}} \\
\hline
\ref{disseminating-false-or-misleading-information}
Disseminating false or misleading information
&

\cite{Wangetal2019, Allcottetal2019, Krittanawongetal2020}

\emph{\cite{Rae2021,Linetal2021, Zhangetal2021, Gwernnet2020, Dale2021, Lacker2020}}
\\
\hline
\ref{causing-material-harm-by-disseminating-false-or-poor-information-e.g.-in-medicine-or-law}
Causing material harm by disseminating misinformation e.g. in medicine
or law
&

\emph{\cite{QuachfortheRegister2020}}

\\
\hline
\ref{leading-users-to-perform-unethical-or-illegal-actions}
Leading users to perform unethical or illegal actions
&
\emph{\cite{Hendrycksetal2020}}
\\
\hline
\multicolumn{2}{@{}>{\raggedright\arraybackslash}l}{}
\\
\multicolumn{2}{@{}>{\raggedright\arraybackslash}l}{
\protect\hyperlink{iv.-malicious-uses}{\color{black}\textbf{Malicious Uses}}} \\
\hline
\ref{making-disinformation-cheaper-and-more-effective}
Making disinformation cheaper and more effective
&

\cite{Zellersetal2020, SchneierforTheAtlantic2020, Hampton2021, Mann2020}
\emph{\cite{HaoforMITTechReview2020, McGuffieNewhouse2020}}

\\
\hline
\ref{facilitating-fraud-scams-and-more-targeted-manipulation}
Facilitating fraud and impersonation scams
&

\cite{Lewisetal2017, VanNoorden2014}
\\
\hline
\ref{assisting-code-generation-for-cyber-attacks-weapons-or-malicious-use}
Assisting code generation for cyber attacks, weapons, or malicious use
&

\emph{\cite{Chenetal2021}}
\\
\hline
\ref{illegitimate-surveillance-and-censorship}
Illegitimate surveillance and censorship
&

\cite{FreedomHouse2019}
\\
\hline
\multicolumn{2}{@{}>{\raggedright\arraybackslash}l}{}
\\
\multicolumn{2}{@{}>{\raggedright\arraybackslash}l}{
\protect\hyperlink{v.-human-computer-interaction-harms}{\color{black}\textbf{Human-Computer Interaction Harms}}} \\
\hline
\ref{anthropomorphising-systems-can-lead-to-overreliance-or-unsafe-use}
Anthropomorphising systems can lead to overreliance or unsafe use
&

\cite{KimSundar2012}
\\
\hline
\ref{create-avenues-for-exploiting-user-trust-nudging-or-manipulation}
Create avenues for exploiting user trust to obtain private information
&

\cite{Broecketal2019, Lewisetal2017, Ischenetal2020}
\\
\hline
\ref{promoting-harmful-stereotypes-by-implying-gender-or-ethnic-identity}
Promoting harmful stereotypes by implying gender or ethnic identity
&
\cite{Curryetal2020, Hwangetal2019, Zdenek2007, Marino2014}
\\
\hline
\multicolumn{2}{@{}>{\raggedright\arraybackslash}l}{}
\\
\multicolumn{2}{@{}>{\raggedright\arraybackslash}l}{
\protect\hyperlink{vi.-automation-access-and-environmental-harms}{\color{black}\textbf{Automation, access, and environmental harms}}} \\
\hline
\ref{environmental-harms-from-operating-lms}
Environmental harms from operating LMs
&

\cite{Strubelletal2019}
\emph{\cite{Benderetal2021, Pattersonetal2021}}
\\
\hline
\ref{increasing-inequality-and-negative-effects-on-job-quality}
Increasing inequality and negative effects on job quality
&
\\
\hline
\ref{undermining-creative-economies}
Undermining creative economies
&
\cite{Hsieh2019}
\\
\hline
\ref{disparate-access-to-benefits-due-to-hardware-software-skill-constraints}
Disparate access to benefits due to hardware, software, skill
constraints
&
\emph{\cite{Benderetal2021}}
\\
\hline
\end{longtable}
  
\end{document}